\documentclass[10pt,twocolumn,letterpaper]{article}

\usepackage{wacv}              

\usepackage[accsupp]{axessibility} 
\usepackage{graphicx}
\usepackage{amsmath}
\usepackage{amssymb}
\usepackage{booktabs}
\usepackage[longtable]{multirow}
\usepackage{pifont}
\usepackage{numprint}
\usepackage{mathtools}
\newcommand\Tstrut{\rule{0pt}{2.6ex}}         
\newcommand\Bstrut{\rule[-0.9ex]{0pt}{0pt}}   
\usepackage{rotating}
\usepackage[nolist]{acronym}
\usepackage{pgfplots}
\usepackage{comment}
\usepackage{pdflscape}
\usepackage{breqn}

\definecolor{tb_color_1}{RGB}{245,124,0}
\definecolor{tb_color_2}{RGB}{0,167,247}
\definecolor{tb_color_3}{RGB}{0,190,0}

\usepackage[pagebackref,breaklinks,colorlinks]{hyperref}

\usepackage[capitalize]{cleveref}
\crefname{section}{Sec.}{Secs.}
\Crefname{section}{Section}{Sections}
\Crefname{table}{Table}{Tables}
\crefname{table}{Tab.}{Tabs.}

\def\wacvPaperID{682} 
\def\confName{WACV}
\def\confYear{2025}

\begin{document}
\begin{acronym}
        \acro{tl}[TL]{transfer learning}
        \acro{da}[DA]{domain adaptation}
        \acro{grl}[GRL]{gradient reversal layer}
\end{acronym}

\title{Cross-domain and Cross-dimension Learning for Image-to-Graph Transformers}

\author{\textbf{Alexander H. Berger}\textsuperscript{1}, \textbf{Laurin Lux}\textsuperscript{1,3}, \textbf{Suprosanna Shit}\textsuperscript{1,5}, \textbf{Ivan Ezhov}\textsuperscript{1},\\\textbf{Georgios Kaissis}\textsuperscript{1,3,4}, \textbf{Martin J. Menten}\textsuperscript{1,2,3}, \textbf{Daniel Rueckert}\textsuperscript{1,2,3}, \textbf{Johannes C. Paetzold}\textsuperscript{2,6}\\
\textsuperscript{1} School of Computation, Information and Technology, Technical University of Munich, Germany\\
\textsuperscript{2} Department of Computing, Imperial College London, UK\\
\textsuperscript{3} Munich Center for Machine Learning (MCML), Munich, Germany\\
\textsuperscript{4} Institute for Machine Learning in Biomedical Imaging, Helmholtz Munich, Germany\\
\textsuperscript{5} Department of Quantitative Biomedicine, University of Zurich, Switzerland\\
\textsuperscript{6} Weill Cornell Medicine, Cornell University, New York City, USA\\
{\tt\small a.berger@tum.de, jpaetzold@med.cornell.edu}}

\maketitle
\begin{abstract}
Direct image-to-graph transformation is a challenging task that involves solving object detection and relationship prediction in a single model. Due to this task's complexity, large training datasets are rare in many domains, making the training of deep-learning methods challenging. This data sparsity necessitates transfer learning strategies akin to the state-of-the-art in general computer vision. In this work, we introduce a set of methods enabling cross-domain and cross-dimension learning for image-to-graph transformers. We propose (1) a regularized edge sampling loss to effectively learn object relations in multiple domains with different numbers of edges, (2) a domain adaptation framework for image-to-graph transformers aligning image- and graph-level features from different domains, and (3) a projection function that allows using 2D data for training 3D transformers. We demonstrate our method's utility in cross-domain and cross-dimension experiments, where we utilize labeled data from 2D road networks for simultaneous learning in vastly different target domains. Our method consistently outperforms standard transfer learning and self-supervised pretraining on challenging benchmarks, such as retinal or whole-brain vessel graph extraction.\footnote{Code: \url{github.com/AlexanderHBerger/cross-dim_i2g}}
\end{abstract}
\vspace{-1em}
\section{Introduction}

\begin{figure}[t]
\centerline{\includegraphics[width=0.95\linewidth]{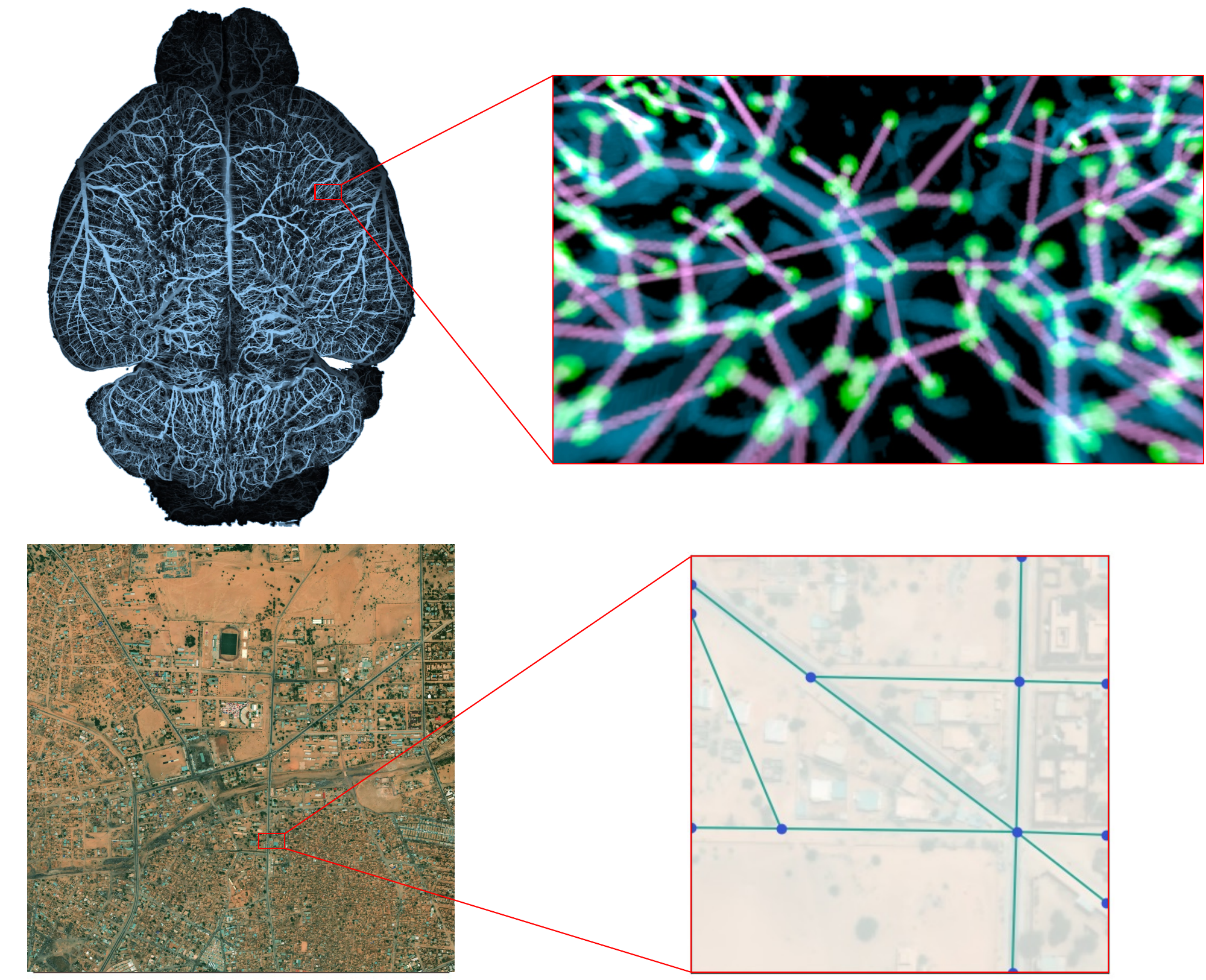}}
\caption{Direct image-to-graph transformation. Whole brain vessel (top) and Agadez road dataset (bottom). The predicted graph is visualized as an overlay on the real image.}
\label{fig:i2g}
\vspace{-0.25cm}
\end{figure}

Representing physical relationships via graph representations has proven to be an efficient and versatile concept with vast utility in machine learning. Prominent examples are road network graphs \cite{bastani2018roadtracer}, neuron representations and connections in the brain \cite{song2019graph_altzheimer}, blood vessels \cite{chen2021retinal}, and cell interactions \cite{zhou2019histology3}. Here, typically used voxelized images disregard the physical structure's semantic content. Hence, constructing graph representations from images (image-to-graph, see Fig. \ref{fig:i2g}) is a critical challenge for unlocking the full potential in many real-world applications \cite{li2022graph}.

\begin{figure*}[t]
\centerline{\includegraphics[width=\linewidth]{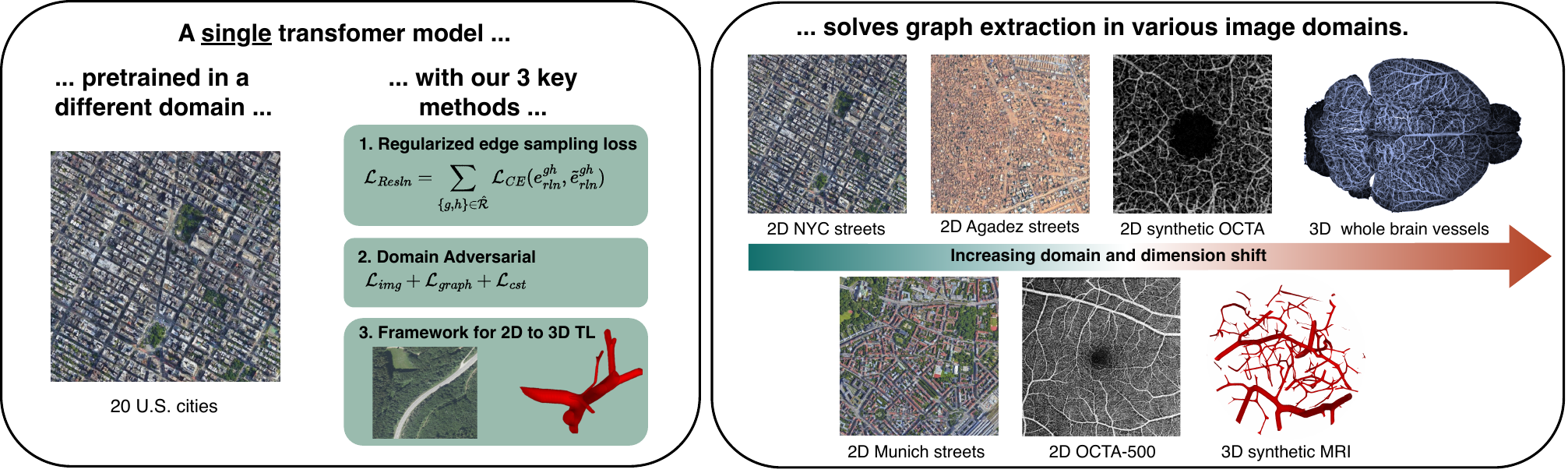}}
\caption{Conceptual overview of our framework. We use a transformer for single-stage image-to-graph transformation. Our three methodological contributions enable knowledge transfer between vastly different domains in 2D and 3D.}
\label{fig:method_overview}
\end{figure*}

Traditionally, image-to-graph transformation involves a complex multi-stage process of segmentation, identification of physical structures and their relations, and iteratively pruning the constructed graph, which leads to inaccuracies and information loss at each step \cite{drees2021voreen}. These disadvantages negatively impact prediction accuracy and limit the application to downstream tasks that require more information from the underlying image (e.g., \cite{sobisch2022automated, shit2022graflow}). Hence, there is a clear need for machine learning solutions that facilitate accurate image-to-graph transformation in a single step \cite{he2020sat2graph,belli2019nodeprune}. Recently, vision transformers have been proposed for this task and showed superior performance to traditional multi-stage approaches \cite{shit2022relationformer, xu2022rngdet++}. However, these approaches require large sets of annotated 2D data and have not been shown to generalize to diverse 3D datasets where graph-level annotations are not widely available.

To address this challenge, we adopt and extend concepts from the field of inductive \ac{tl} (see taxonomy in \cite{pan2009survey}), which have not been explored for image-to-graph transformation. Inductive TL simultaneously uses large annotated datasets in the source domain (e.g., 2D satellite images) and a small set of labeled data in the target domain (e.g., 3D microscopy images of vascular networks). 

\paragraph{Our contribution.}
Guided by the hypothesis that the \textit{underlying graph representations of physical networks are similar across domains}, we introduce a set of methodological innovations for image-to-graph transformation:

\begin{enumerate}
    \item We introduce a loss formulation that regularizes the number of sampled edges during training, allowing simultaneous learning in domains that differ in terms of the number of objects and relations (\cref{sec:loss}).
    \item We propose a supervised domain adaptation framework for image-to-graph transformers that aligns features from different domains (\cref{sec:domain_adaption}).
    \item We propose a framework for training 3D image-to-graph models with 2D data. Our framework introduces a projection function from the source representation to a space similar to the target domain (\cref{sec:tl_framework}). 
\end{enumerate}

\noindent In extensive experiments on six datasets, we demonstrate our method's utility and outperform all baselines, validating our initial hypothesis. Our method leads to \textit{four main results}: 1) we show compelling improvements in image-to-graph transformation on existing datasets; 2) our method enables image-to-graph transformation in previously unsolvable sparse data scenarios; 3) our method outperforms self-supervised pretraining by a margin and 4) our method bridges dimensions during training, i.e. we solve the previously unsolved direct image-to-graph inference for complex 3D vessel images by utilizing 2D road data in an additional training step. Further ablation studies highlight that each of our methodological contributions addresses a specific weakness in the state-of-the-art.

\section{Related Works}

\paragraph{Image-to-graph transformation.}

Image-to-graph transformation is an increasingly important field in computer vision with application to various domains, including road network extraction from satellite images \cite{wang2016roadreview} or 3D vessel graph extraction from microscopy images \cite{batten2023imagetotree, almasi2015graphextractionmedical}. Traditional methods solve this task through a multi-step approach involving input segmentation \cite{chen2022road, bastani2018roadtracer, carion2020detr}, skeletonization, and pruning to generate the graph \cite{wang2016roadreview, drees2021voreen}. Deep learning-based approaches frequently use an object detector followed by relation prediction \cite{batten2023imagetotree, almasi2015graphextractionmedical} or require a segmentation map as input \cite{xu2022rngdet++}. These approaches' performances are determined by the performance of the pipeline's intermediate stages. Furthermore, information loss at each stage limits performance and applicability to downstream tasks. Lastly, these methods are tailored to specific domains, rendering them unsuitable for cross-domain applications, including \ac{tl}.

We select the \textit{Relationformer} as our base concept because it is the most general single-stage transformer concept that can directly predict the graph from diverse image data in various domains \cite{shit2022relationformer, prabhakar2023vesselformer}. This generalizability makes it especially suitable for cross-domain \ac{tl}.

\paragraph{Transfer learning for transformers.} Recent studies showed effective pretraining for transformers on natural images. Prevailing architectures involve supervised pretraining of the model's backbone on, e.g., ImageNet, coupled with random initialization of the encoder-decoder component \cite{shit2022relationformer, ma2023revisiting, carion2020detr}. Dai et al. \cite{Dai2021upDETR} extend DETR \cite{carion2020detr} with a specific pretext task for object detection. Li et al. \cite{Li2021transferViT} compared self- and supervised pretraining methods with random initialization of a ViT-backbone \cite{dosovitskiy2020vit}. They showed how self-supervised pretraining improves downstream object detection without a specific pretext task. Ma et al. \cite{ma2023revisiting} pretrained a transformer architecture on synthetic data and outperformed self-supervised methods for object detection. However, generating synthetic data, especially for specialized tasks, is typically domain-specific or requires expert knowledge \cite{kreitner2023detailed}. Ma et al. \cite{ma2023revisiting} find that random weight initialization remains a robust baseline, often achieving competitive downstream performance. To this date, no study has explored \ac{tl} for image-to-graph transformers.

\paragraph{Cross-domain transfer learning.}
Existing cross-domain \ac{tl} approaches are generative or discriminative. Generative approaches translate images from source to target domain on a pixel-level using a generative network \cite{zhu2017unpaired}. In restricted settings, they have shown promising results, e.g., object detection for day-/nighttime shifts in road scenes \cite{huang2018auggan} or MRI to CT translation \cite{hammami2020cycle}. However, generative approaches require additional training of the translation network, which is computationally expensive and suffers from training instability \cite{lucic2018gans}. Discriminative approaches train a single model to learn a general representation that is transferable between domains \cite{wang2018deep}. Some utilize a domain adversarial network that distinguishes whether a sample is from the source or target domain based on its feature representation. Combined with a \ac{grl} \cite{ganin2015unsupervised}, this approach has proven effective in classification \cite{ganin2015unsupervised}, segmentation \cite{wang2020classes}, and object detection \cite{chen2018domain}. Chen et al. \cite{chen2018domain} introduced a \ac{da} framework for object detection using the \ac{grl} concept with two domain classifiers at the image and instance levels. Both are based on $\mathcal{H}$-divergence theory \cite{ben2010theory}. To reduce the bias towards the source domain, a consistency regularization penalizes the $L_2$-distance between the domain classifiers \cite{chen2018domain}. To this day, all existing approaches are limited to a relatively small domain shift (e.g., from day to night scenes or synthetic to real images) instead of a \textit{fundamental} domain shift, such as from satellite images to medical scans. Also, none of the existing approaches have been applied to the image-to-graph transformation problem.

\paragraph{Cross-dimension transfer learning.}
2D to 3D \ac{tl} is a highly challenging but promising research direction because of the abundance of labeled 2D data compared to the scarcity of 3D data. The few existing approaches \cite{shen2022simcrosstrans, liu2021pixeltopoint, xie2021unified2d3d, wang2022vision2dto3d} address the challenge, either data-based or model-based. Data-based approaches augment and project 2D data into 3D space, while model-based approaches aim to adjust the model to work with multi-dimensional input. Shen et al. \cite{shen2022simcrosstrans} introduced a method that projects 3D point clouds to pseudo-2D RGB images. Liu et al. \cite{liu2021pixeltopoint} introduced a pixel-to-point knowledge transfer that pretrains a 3D model by generating 3D point cloud data from 2D images with a learned projection function. A model-based approach by Xie et al. \cite{xie2021unified2d3d} used dimension-specific feature extractors and a dimension-independent Transformer. Similarly, Wang et al. \cite{wang2022vision2dto3d} proposed a special tokenizer that creates 2D-shaped patch embeddings in a standard 2D ViT-model. 
All existing approaches either require changes to the target model, which may limit performance, a specifically crafted projection function, or additional training (e.g., for learning an explicit projection function). In our approach, we seek simplicity in implementation and training as well as generalizability to new domains and tasks.

\section{Methodology}
In this section, we describe our three key contributions to efficiently transfer knowledge from a source domain with image and graph space denoted as a $(\mathcal{I}^S, \mathcal{G}^S)$ to a target domain with image and graph space denoted as $(\mathcal{I}^T, \mathcal{G}^T)$ for an image-to-graph transformer.

\subsection{Regularized Edge Sampling Loss}
\label{sec:loss}

A leading difference between the source and target domain in our cross-domain \ac{tl} setting is the different node and edge distribution, which poses a challenge because they dictate the relation loss calculation. 
Previous works arbitrarily pick the number of ground truth edges and fix it for the relation loss calculation ($\mathcal{L}_{\mathrm{rln}}$) for a specific dataset \cite{shit2022relationformer}, which evidently generalizes poorly across varying domains. Generally, such a loss originates from object detection, where pair-wise relations are classified with a cross-entropy (CE) loss over a fixed number of edges $m$. This includes all active edges $\mathcal{A}$ and an irregular number of randomly sampled background edges $\bar{\mathcal{B}} \subseteq \mathcal{B}$. An active edge is defined as a pair of nodes ${g,h}$ that is connected by an edge, i.e., $e^{gh}_{rln}=1$. Similarly, for a background edge, $e^{gh}_{\mathrm{rln}}=0$ holds. Then, $\mathcal{L}_{\mathrm{rln}}$ is the CE loss on the set of sampled edges $\mathcal{R}= \mathcal{A} \cup \bar{\mathcal{B}}$ with $|\mathcal{A}| + |\bar{\mathcal{B}}| \leq m$:
\begin{equation}
    \mathcal{L}_{\mathrm{rln}}=\sum_{\{g,h\}\in\mathcal{R}}{\mathcal{L}_{\mathrm{CE}}(e_{\mathrm{rln}}^{gh},\tilde{e}_{\mathrm{rln}}^{gh})}
\end{equation}
where $\tilde{e}^{gh}_{\mathrm{rln}} \in \{0,1\}$ is the model's relation prediction for the respective pair of nodes. Crucially, this formulation ignores the subset $\mathcal{B} \setminus \bar{\mathcal{B}} $ when calculating the loss. 

In previous works \cite{shit2022relationformer}, $m$ is a manually chosen global hyperparameter, strongly affecting the model's performance. If $m$ is too small, not enough background edges are sampled; hence, the loss does not penalize over-prediction. If $m$ is too large, background edges dominate the edge loss calculation because the edge space is sparse. In that case, the network under-predicts edges. Furthermore, with only a small subset of all edges being sampled, the loss gives a noisy signal regarding edge prediction, which worsens the learning process overall.

To address these limitations, we introduce our regularized edge sampling loss for transformers, short $\mathcal{L}_{\mathrm{Reslt}}$. In simple terms, $\mathcal{L}_{\mathrm{Reslt}}$ adaptively chooses the number of sampled edges. If necessary, $\mathcal{L}_{\mathrm{Reslt}}$ upsamples the edges up to a fixed ratio between active and background edges. 
With our novel approach, we achieve a consistent loss across samples from different domains. Formally, we introduce our regularized edge sampling below:

$\hat{\mathcal{R}} = \hat {\mathcal A}\cup \hat {\mathcal B}$ is the set of a batch's upsampled edges. The number of elements in the upsampled multisets $\mathcal {\hat A}$ and $\mathcal {\hat B}$ have the pre-defined ratio $r=\frac {|\mathcal {\hat A}|} {|\mathcal {\hat B}|}$ where $r \in [0,1]$. Multisets are necessary because the ratio is achieved by duplicating random edges in $\mathcal{A}$ or $\mathcal{B}$. 

A batch's labels consist of a ground-truth graph $\mathcal{G}_n$ for each sample $n$. $\mathcal{G}_n$ is defined as a tuple of the sample's nodes and edges, $\mathcal{G}_n=(\mathcal{V}_n,\mathcal{E}_n)$.
The set of a batch’s nodes $\mathcal{V}$ and edges $\mathcal{E}$ is thus defined as:
\begin{equation}
\mathcal{V}=\bigcup_{n=0}^{N}\mathcal{V}_n
    \quad \text{and}  \quad
    \mathcal{E} = \bigcup_{n=0}^{N}\mathcal{E}_n
\end{equation}
Each active edge $a \in \mathcal{A} = \mathcal{E}$ is a tuple of two nodes $(g,h)$ that are connected by the respective edge:
\begin{equation}
    a=(g,h) \in \mathcal{E} \quad \text{where}  
\quad g,h \in \mathcal{V}
\end{equation}
Each background edge $b \in \mathcal{B}$ is a tuple of two nodes $(j,k)$ that are not connected by an edge:
\begin{equation}
    b=(j,k) \in (\mathcal{V} \times \mathcal{V}) \setminus \mathcal{E}=\mathcal B \quad
\end{equation}
Then, the upsampled multiset of active edges $\mathcal {\hat A}$ is:
\begin{equation}
\label{eq:frl_cond_active}
\mathcal {\hat A} = \mathcal{E} \cup 
[x_i | x_i = a_{i \bmod |\mathcal{E}|}]
\end{equation}
with $
|\mathcal{E}| \leq i < |\mathcal B| * r \quad \text{and} \quad i \in \mathbb{N}.~ 
$
Similarly, we define the upsampled multiset of background edges $\mathcal {\hat B}$ as:
\begin{equation}
\label{eq:frl_cond_background}
    \mathcal {\hat B} = \mathcal B
    \cup [x_i | x_i = b_{i \bmod |\mathcal B|}]
\end{equation}
with:
$
    |\mathcal B| \leq i < \frac {|\mathcal{E}|}{r} 
$.
Notably, only one of the sets is upsampled while the other stays the same because only one of the conditions in \cref{eq:frl_cond_active} or \cref{eq:frl_cond_background} can produce a valid $i$.

Hence, we define our regularized edge sampling loss as:
\begin{equation}
    \mathcal{L}_{\mathrm{Reslt}}=\sum_{\{g,h\}\in \hat{\mathcal{R}}}{\mathcal{L}_{\mathrm{CE}}(e_{\mathrm{rln}}^{gh},\tilde{e}_{\mathrm{rln}}^{gh})}
\end{equation}

Although the edge ratio $r$ is a hyperparameter, the model's performance is relatively insensitive to its value. A default value of $0.15$ showed good results across all datasets (see \cref{sec:results} and Supplement), which makes $\mathcal{L}_{\mathrm{Reslt}}$ beneficial compared to the previous works' stochastic edge losses. Furthermore, $\mathcal{L}_{\mathrm{Reslt}}$ gives a precise signal regarding edge prediction and increases convergence speed.

\subsection{Supervised Domain Adaptation}
\label{sec:domain_adaption}
In our setting, the stark differences between the source and target domain in image and graph features further amplify the \ac{tl} challenge. Specifically, source and target domains significantly differ in image characteristics such as background and foreground intensities, signal-to-noise ratio, and background noise, as well as graph characteristics such as the structures' radii or edge regularity. Edge regularity refers to the geometrical straightness of the underlying structure. While roads in the U.S. (e.g., highways) typically have a high edge regularity, vessels in microscopic images are highly irregular (i.e., the vessel does not follow a straight line and has high intra-edge curvature).
To address this challenge, we utilize a domain adversarial network on the image and graph level, respectively. These adversarial networks are used when jointly pretraining in both domains. Similar to previous methods \cite{ganin2015unsupervised, tzeng2017adversarial, long2018conditional}, the image-level adversarial network is a small neural network that classifies the domain based on a sample's feature representation after the feature extractor. We treat each image patch $p^{u,v}$ at position $u,v$ as an individual sample and compute the CE loss 
\begin{equation}
    \mathcal{L}_{\mathrm{img}} = - \sum_{u,v} \Bigl[ D \text{ log } p^{u,v} + (1 - D) \text{ log } (1 - p^{u,v})\Bigr]
\end{equation}
where $D \in \{0,1\}$ denotes whether the respective sample is from the source or target domain \cite{chen2018domain}. 

To align graph-level features, we view the concatenated tokenized transformer output $T \in 	\mathbb{R}^{(\#o + \#r) \times d}$ as a sample's abstract graph representation where $\#o$ and $\#r$ are the numbers of object and relation tokens, respectively, and $d$ is the amount of the tokens' hidden channels. We train a domain classifier on this abstract graph representation using the CE loss:
\begin{equation}
    \mathcal{L}_{\mathrm{graph}} = - \Bigl[ D \text{ log } T + (1 - D) \text{ log } (1 - T)\Bigr]
\end{equation}
Both domain classifiers are preceded by a \ac{grl} \cite{ganin2015unsupervised} reversing the gradient such that the main network is learning to maximize the domain loss. This framework forces the network to learn domain invariant representations and, thus, aligns the source and target domain. Furthermore, we apply a consistency regularization between both domain classifications to reduce the bias towards the source domain, as shown in \cite{chen2018domain}. This consistency regularization is expressed by an objective function minimizing the $L_2$-distance between the domain classifiers' predictions (i.e., the output of the classification functions $d_{\mathrm{img}}$ and $d_{\mathrm{graph}}$):
\begin{equation}
    \mathcal{L}_{\mathrm{cst}} = \Bigl|\Bigl| \frac{1}{|I|}\sum_{u,v} d_{img}(p^{u,v}) - d_{graph}(D) \Bigr|\Bigr|_2
\end{equation}
where we take the average over the image-level patch classifications of a sample consisting of $|I|$ patches.

\subsection{Combined Training Loss}
\label{sec:complete_loss_formulation}

Our new regularized edge loss is combined with the other essential loss components to a final optimization function. The final loss consists of the $L_1$ regression loss ($\mathcal{L}_{\mathrm{reg}}$), the scale-invariant, generalized intersection over union loss ($\mathcal{L}_{\mathrm{gIoU}}$) for the box predictions (with predicted boxes $\tilde{v}_{box}$ and ground truth $v_{\mathrm{box}}$), and a CE classification loss ($\mathcal{L}_{\mathrm{cls}}$) for object classification \cite{shit2022relationformer}. Further, our new regularized edge sampling loss  $\mathcal{L}_{\mathrm{Reslt}}$ and our three \ac{da} losses ($\mathcal{L}_{\mathrm{img}}$, $\mathcal{L}_{\mathrm{graph}}$, and $\mathcal{L}_{\mathrm{cst}}$) are included. Furthermore, in order to achieve unique predictions, we compute a bipartite matching between the ground truth and predicted objects utilizing the Hungarian algorithm \cite{shit2022relationformer}. $\mathcal{L}_{\mathrm{reg}}$, $\mathcal{L}_{\mathrm{gIoU}}$, and $\mathcal{L}_{\mathrm{Reslt}}$ are calculated over all object predictions $v$ that are matched to a ground truth, i.e. where $v_{\mathrm{cls}}^{i} = 1$, whereas $\mathcal{L}_{\mathrm{cls}}$ is calculated over all object predictions. The combined loss term for our $N$ object tokens in a batch is defined as:

\begin{equation}
\begin{aligned}
    \mathcal{L} = & \sum^N_{{i=1},\ {[v_{\mathrm{cls}}^{i}} = 1]} \Bigl[ \lambda_{\mathrm{reg}} \mathcal{L}_{\mathrm{reg}}(v^i_{\mathrm{box}},\tilde{v}^i_{\mathrm{box}}) \\
    & \quad \quad \quad \quad+ \lambda_{\mathrm{gIoU}} \mathcal{L}_{\mathrm{gIoU}}(v^i_{\mathrm{box}},\tilde{v}^i_{\mathrm{box}}) \Bigr] \\
    &+ \lambda_{\mathrm{cls}} \sum_{i=1}^{N}{\mathcal{L}_{\mathrm{cls}}(v^i_{\mathrm{box}},\tilde{v}^i_{\mathrm{box}})} \\
    &+ \lambda_{\mathrm{DA}} (\mathcal{L}_{\mathrm{img}} + \mathcal{L}_{\mathrm{graph}} + \mathcal{L}_{\mathrm{cst}})\\
    &+ \lambda_{\mathrm{Reslt}} \underbrace{\sum_{\{ g, h \} \in \hat{\mathcal{R}}}{\mathcal{L}_{\mathrm{CE}}(e^{gh}_{\mathrm{rln}},\tilde{e}^{gh}_{\mathrm{rln}})}}_{\mathcal{L}_{\mathrm{Reslt}}}
\end{aligned}
\end{equation}

with $\lambda_{\mathrm{reg}}$, $\lambda_{\mathrm{gIoU}}$, $\lambda_{\mathrm{cls}}$, $\lambda_{\mathrm{Reslt}}$, and $\lambda_{\mathrm{DA}}$ as weights.

\subsection{Framework for 2D-to-3D Transfer Learning}
\label{sec:tl_framework}
This section describes our framework for the challenging setting of a 2D source domain and a 3D target domain. This setting is especially relevant given the scarcity of completely annotated 3D image datasets. At the core of our framework is a simple projection function $\Pi$ that transforms source instances into a space similar to the target space, i.e., $\Pi: (\mathcal{I}^S, \mathcal{G}^S) \longrightarrow (\bar{\mathcal{I}}, \bar{\mathcal{G}})$. Since our regularized edge sampling loss (\cref{sec:loss}) and domain adaptation framework (\cref{sec:domain_adaption}) automatically optimize the alignment of source and target domain characteristics, we do not need to engineer our projection to resemble the target domain characteristics (e.g., signal-to-noise ratio or the structures' radiuses). Thus, we can design our projection function in the most simple and generalizable form. Intuitively,  $\Pi$ projects 2D data to a 3D space by simply creating an empty 3D volume, placing the 2D image as a frame in it, and randomly rotating the entire volume. Formally, $\Pi$ is described by:

\begin{enumerate}
    \setlength\itemsep{0cm}
    \item Resize $\mathcal{I}^S$ from $(H^{\mathcal{I}^S} \times W^{\mathcal{I}^S})$ to the target domain's spatial patch size ($H^T \times W^T$) by a linear downsampling operator $D:{\mathcal{I}^S}\rightarrow {\mathcal{I}^S}^\prime$, where $D\in \mathbb{R}^{H^TW^T\times H^{\mathcal{I}^S} W^{\mathcal{I}^S}}$. $G$ remains unchanged as we use normalized coordinates.
    \item We initialize $I$ in 3D with $I = \boldsymbol{0}^{H^T W^T D^T}$ and place ${\mathcal{I}^S}^\prime$ in $I$ at slice location $z=0.5$. We also augment the node coordinates of $G:={V,E}$ by $V^\prime=\{[v, 0.5]:v\in V\}$. New graph $G^\prime:=(V^\prime,E)$.
    \item We apply a random three dimensional rotation matrix $R$ on $I$ and obtain $\bar{I} \in \bar{\mathcal{I}}$.  We apply the same $R$ on the nodes of $G^\prime$ and obtain $V^{\prime\prime}=\{Rv:v\in V\}$. New graph $\bar{G}:=(V^{\prime\prime},E) \in \bar{\mathcal{G}}$.
\end{enumerate}
Notably, our approach works out of the box without requiring segmentation masks, handcrafted augmentations, specifically engineered projections, or changes to the target model. Furthermore, it naturally extends to new domains and is trainable end-to-end together with the target task.
\section{Experiments and Results}
\label{sec:results}
\paragraph{Datasets.}
We validate our method on a diverse set of six public image datasets capturing physical networks. 
We choose two 2D road datasets, namely a dataset from Munich (a European city with green vegetation-dominated land cover) and from Agadez (a historic Tuareg city in Niger in the Sahara desert). The appearance in satellite images of these cities and their network structure substantially differ from the pretraining set as well as from each other; see \cref{fig:method_overview}. Accurately extracting road graphs is a highly important task for traffic forecasting and traffic flow modeling \cite{ermagun2018spatiotemporal,martin2020graph}. Next, we choose a synthetic OCTA retina dataset \cite{menten2022physiology} and a real OCTA dataset \cite{li2020octa500}. Additionally, we present experiments on two 3D datasets, namely a synthetic vessel dataset \cite{schneider2012tissue} and a real whole-brain microscopy vessel dataset \cite{todorov2020machine}. Details on the datasets and data generation can be found in the Supplement.

\paragraph{Training.} We pretrain our method on the 20 U.S. cities dataset \cite{he2020sat2graph} jointly with the target dataset. We crop the source images to overlapping patches, in which we eliminate redundant nodes (i.e., nodes of degree 2 with a curvature of fewer than 160 degrees) to train our model on meaningful nodes \cite{belli2019nodeprune}. After pretraining, we finetune the model on the target dataset for 100 epochs. For more details, please refer to the Supplement and our repository.

\begin{table*}[t]
\centering
\footnotesize
\caption{Main results. Quantitative Results for our cross-domain and cross-dimensional image-to-graph transfer learning framework. The domain shift increases from top to bottom. We outperform the baselines across all datasets. The best scores per respective metric across all models for a dataset are highlighted in bold. $\star$ Results for supervised pretraining on D) and E) (3D data) are not reported because it is technically not possible. Additional metrics and standard deviations are given in the Supplement.}
\begin{tabular}{ l | l | c | c | c | c | c | c | c }
    \toprule
    \begin{tabular}[c]{@{}l@{}} \textbf{Fine Tuning} \\ \textbf{Training Set}\end{tabular} 
    & \begin{tabular}[c]{@{}l@{}} \textbf{(Pre-)Training} \\ \textbf{Strategy}\end{tabular} 
    & \begin{tabular}[c]{@{}l@{}} \textbf{Node-}     \\ \textbf{mAP}$\uparrow$\end{tabular} 
    & \begin{tabular}[c]{@{}l@{}} \textbf{Node-}     \\ \textbf{mAR}$\uparrow$\end{tabular} 
    & \begin{tabular}[c]{@{}l@{}} \textbf{Edge-}     \\ \textbf{mAP}$\uparrow$\end{tabular} 
    & \begin{tabular}[c]{@{}l@{}} \textbf{Edge-}     \\ \textbf{mAR}$\uparrow$\end{tabular} 
    & \textbf{SMD $\downarrow$}
        & \begin{tabular}[c]{@{}l@{}} \textbf{Topo-} \\ \textbf{Prec.$\uparrow$ }\end{tabular} 
    & \begin{tabular}[c]{@{}l@{}} \textbf{Topo-} \\ \textbf{Rec.$\uparrow$ }\end{tabular} \\

    \hline
    \midrule 
    \multicolumn{4}{l}{\textbf{A) TL from roads (2D) to roads (2D)} \Tstrut\Bstrut} \\
    \midrule
    \multirow{4}{*}{Agadez \cite{haklay2008osm}}
    &  No Pretraining \cite{glorot2010understanding} & 0.067 & 0.122 & 0.021 & 0.043 & 0.062 & 0.369 & 0.261\\
    &  Self-supervised \cite{chen2021empirical}      & 0.083 & 0.156 & 0.030 & 0.071 & 0.030 & 0.471 & 0.459\\
    &  Supervised                                    & 0.161 & 0.237 & 0.115 & \textbf{0.177} & 0.023 & 0.783 & \textbf{0.711}\\
    &  \textbf{Ours}                                 & \textbf{0.163} & \textbf{0.244} & \textbf{0.116} & 0.172 & \textbf{0.022} & \textbf{0.816} & 0.614\\
    \hline
    \multirow{4}{*}{Munich \cite{haklay2008osm}}
    &  No Pretraining \cite{glorot2010understanding} & 0.083 & 0.120 & 0.034 & 0.054 & 0.235 & 0.260 & 0.247\\
    &  Self-supervised \cite{chen2021empirical}      & 0.088 & 0.145 & 0.060 & 0.097 & 0.155 & 0.339 & 0.384\\
    &  Supervised                                    & 0.277 & 0.336 & 0.207 & 0.272 & 0.091 & 0.682 & \textbf{0.660}\\
    &  \textbf{Ours}                                 & \textbf{0.285} & \textbf{0.344} & \textbf{0.224} & \textbf{0.277} & \textbf{0.090} & \textbf{0.726} & 0.655\\
    \hline
    \midrule
    \multicolumn{9}{l}{\textbf{B) TL from roads (2D) to synthetic retinal vessels (2D) }\Tstrut\Bstrut}  \\
    \midrule
    \multirow{4}{*}{\begin{tabular}[c]{@{}l@{}} Synthetic \\ OCTA \cite{menten2022physiology}\end{tabular}}
    &  No Pretraining \cite{glorot2010understanding} & 0.273 & 0.375 & 0.140 & 0.339 & 0.005 & 0.181 & 0.948\\
    &  Self-supervised \cite{chen2021empirical}      & 0.136 & 0.260 & 0.069 & 0.223 & 0.031 & 0.093 & 0.927\\
    &  Supervised                                    & 0.291 & 0.384 & 0.170 & 0.338 & 0.004 & 0.211 & \textbf{0.957}\\
    &  \textbf{Ours}                                 & \textbf{0.415} & \textbf{0.493} & \textbf{0.250} & \textbf{0.415} & \textbf{0.002} & \textbf{0.401} & 0.890\\
    \hline
    \midrule
    \multicolumn{9}{l}{\textbf{C) TL from roads (2D) to real retinal vessels (2D) }\Tstrut\Bstrut}  \\
    \midrule
    \multirow{4}{*}{ OCTA-500 \cite{li2020octa500} }
    &  No Pretraining \cite{glorot2010understanding} & 0.189 & 0.282 & 0.108 & 0.169 & 0.017 & 0.737 & 0.634\\
    &  Self-supervised \cite{chen2021empirical}      & 0.214 & 0.305 & 0.135 & 0.213 & 0.016 & 0.763 & 0.706\\
    &  Supervised                                    & 0.366 & 0.447 & 0.276 & 0.354 & 0.014 & 0.862 & 0.775\\
    &  \textbf{Ours}                                 & \textbf{0.491} & \textbf{0.571} & \textbf{0.366} & \textbf{0.489} & \textbf{0.012} & \textbf{0.877} & \textbf{0.817}\\
    \hline
    \midrule
    \multicolumn{9}{l}{\textbf{D) TL from roads (2D) to brain vessels (3D)}\Tstrut\Bstrut} \\
    \midrule
    \multirow{4}{*}{\begin{tabular}[c]{@{}l@{}} Synthetic \\ MRI \cite{schneider2012tissue}\end{tabular}}
    &  No Pretraining \cite{glorot2010understanding} & 0.162 & 0.250 & 0.125 & 0.201 & \textbf{0.013} & - & -\\
    &  Self-supervised \cite{chen2021empirical}      & 0.162 & 0.252 & 0.120 & 0.193 & 0.014 & - & -\\
    &  Supervised                                    & $\star$ & $\star$ & $\star$  & $\star$  & $\star$  & $\star$ & $\star$\\
    &  \textbf{Ours}                                 & \textbf{0.356} & \textbf{0.450} & \textbf{0.221} & \textbf{0.322} & \textbf{0.013} &  - & -\\
    \hline
    \midrule
    \multicolumn{9}{l}{\textbf{E) TL from roads (2D) to real whole-brain vessel data (3D)}\Tstrut\Bstrut} \\
    \midrule
    \multirow{4}{*}{\begin{tabular}[c]{@{}l@{}} Microscopic \\ images \cite{todorov2020machine}\end{tabular}}
    &  No Pretraining \cite{glorot2010understanding} & 0.231 & 0.308 & 0.249 & 0.329 & \textbf{0.017} & - & -\\
    &  Self-supervised \cite{chen2021empirical}      & 0.344 & 0.404 & 0.363 & 0.425 & \textbf{0.017} & - & -\\
    &  Supervised                                    & $\star$  & $\star$  & $\star$  & $\star$  & $\star$  & $\star$ & $\star$\\
    &  \textbf{Ours}                                 & \textbf{0.483} & \textbf{0.535} & \textbf{0.523} & \textbf{0.566} & \textbf{0.017} & - & -\\
    \bottomrule
\end{tabular}
\label{table:main_results}
\vspace{-0.3cm}
\end{table*}

\paragraph{Metrics.}
We evaluate our method on six metrics from object detection and graph similarity tasks. For graph similarity, we report the 2D TOPO-score \cite{biagioni2012topo} and the street mover distance (SMD), which approximates the Wasserstein distance of the graph \cite{belli2019nodeprune}. From object detection, we report mean average recall (mAR) and mean average precision (mAP) for node- and edge-detection. For more implementation details, please refer to the Supplement.

\paragraph{Baselines.}
No prior work has developed transfer learning techniques for the structural image-to-graph transformation problem. To evaluate the significance of our proposed methods, we compare the downstream task performance against three competing approaches with varying pretraining and initialization methods. Our first baseline, \textit{no pretraining}, is random weight initialization \cite{glorot2010understanding}, which is considered standard practice for model initialization when no suitable pretraining is available. Second, we benchmark against a state-of-the-art method for \textit{self-supervised pretraining}, MoCo v3 \cite{chen2021empirical}, which even outperformed supervised pretraining in some tasks \cite{chen2021empirical}. Self-supervised pretraining is typically used when the amount of unlabeled data significantly exceeds that of labeled data in the same (or very similar) domain. Hence, we pretrain on a large set of unlabeled data from the same domain in each experiment for the self-supervised baseline. For more details regarding the unlabeled dataset, please refer to the Supplement. Third, \textit{supervised pretraining}, where we pretrain the target model on the source data without using our methodological contributions. This approach has been successfully applied for various vision transformers, including the relationformer \cite{shit2022relationformer}. Note that the \textit{supervised pretraining} baseline is impossible in 3D scenarios; only a projection function enables the use of 2D data for pretraining 3D models.

\subsection{Results on Cross-domain TL (2D)}
\label{sec:results_roads}

Our proposed transfer learning strategy shows excellent results across 2D datasets. We outperform the baseline without pretraining and self-supervised pretraining on all datasets across all object detection and graph similarity metrics; see \cref{table:main_results}. As the domain shift increases, we significantly outperform naive pretraining. 

\begin{figure}[t!]
\centerline{\includegraphics[width=\linewidth]{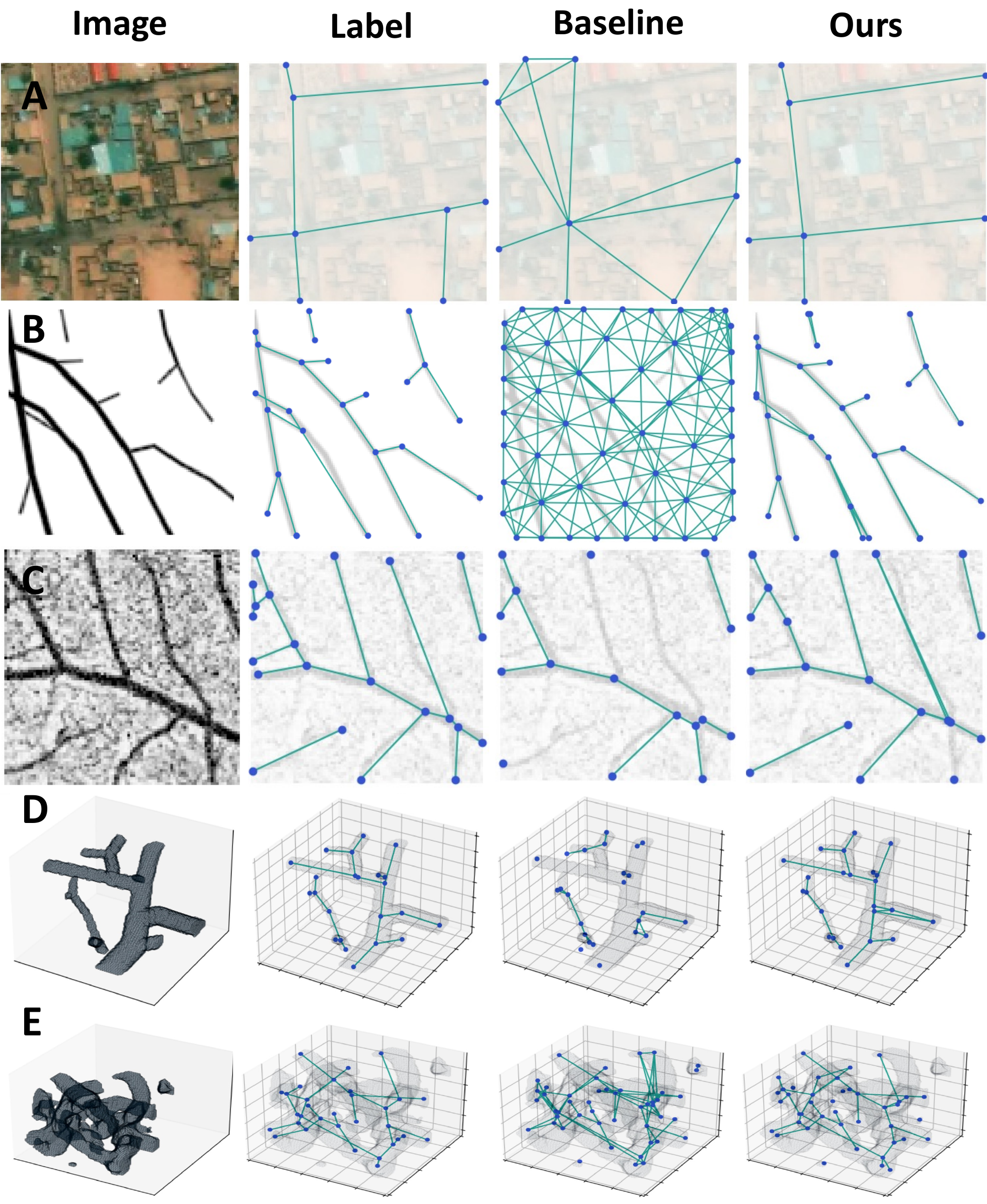}}
\caption{Qualitative results. From left to right: Image, ground truth graph, \textit{no pretraining}-baseline, and our method. Datasets in each row are indicated by the letters for the datasets as in \cref{table:main_results}. Our method consistently outperforms the \textit{no pretraining} baseline, which overpredicts the edges and nodes in all datasets but the OCTA-500, where the fine-tuning set is uncharacteristically large.}
\label{fig:qual_results}
\end{figure}

\paragraph{Roads in diverse locations.}  
First, we show that we can learn to extract road graphs in topographically diverse locations with vastly variant land cover via TL. On both datasets (see \cref{table:main_results} A), we tripled the performance across almost all metrics compared to our baseline. Our results show that edge detection fails without any form of transfer learning. Although the self-supervised method improves the baseline across all metrics, the performance does not reach the level of supervised pretraining. While our approach yields the best performance, the difference to the baseline with naive pretraining is small. We attribute this small difference to the small domain shift between the source (U.S. roads) and the target domains
, which eases knowledge transfer.

\paragraph{Retinal blood vessels.}
In the next experiment, we introduce a larger domain shift in our TL. Our target sets are two retinal blood vessel datasets (see \cref{table:main_results} B and C). Our method doubles the node and edge detection performance on the OCTA-500 dataset \cite{li2020octa500} compared to "no pretraining." Furthermore, we significantly increase object detection and graph similarity metrics across both datasets compared to all baselines. The qualitative examples (\cref{fig:qual_results}) indicate that this improvement is associated with identifying more correct nodes and edges. We observe that the self-supervised method does not improve performance on the synthetic dataset but still achieves minor improvements for OCTA-500. We attribute this to the differences between the target sets: in the OCTA-500, only the arterioles and venules are annotated, leading to an easy topological structure. The main difficulty here lies in differentiating foreground from background, a task in which contrastive self-supervised training shows excellent performance. While this differentiation is easy in the synthetic OCTA dataset, the main difficulty is learning the topological structure with many (often overlapping) edges (see \cref{fig:qual_results} and the Supplement). Learning this complex structure requires label information, as visible in the superior performance of the supervised pretraining methods. Naive pretraining still improves performance, but we observe a large (compared to A) performance difference between naive pretraining and our method. We attribute this to the larger domain shift, which is better addressed by our proposed methodology.

\subsection{Results on Cross-dimension TL (2D to 3D)}
Finally, we explore dimensional shifts in addition to a stark domain shift. Leveraging our new proposed loss (\cref{sec:loss}), our \ac{da} framework (\cref{sec:domain_adaption}), and our 2D-3D projection function (\cref{sec:tl_framework}), we pretrain models on raw satellite images for the challenging task of 3D vessel graph extraction on a synthetic and a real dataset. Our experiments on the VesSAP dataset \cite{todorov2020machine} show strong improvements in all graph similarity and object detection scores (see \cref{table:main_results} E). The self-supervised method also displays improvements, which, however, do not reach our method's performance. Similarly, our method roughly doubles the object detection metrics on the 3D MRI dataset compared to the baselines (see \cref{table:main_results} D).  

We do not report results for naive pretraining because, in contrast to our method, it simply does not allow for 2D-to-3D TL. Our ablation in \cref{sec:ablation} shows results for applying our projection function only, without any use of our other proposed contributions. When studying the lower-performing baselines, we observe that self-supervised pretraining with MoCo v3 leads to higher improvements in the real microscopic vessel data compared to the synthetic MRI dataset. Both datasets have complex topologies that require supervised training (see \cref{sec:results_roads} ), but only the real dataset has high-intensity variations, which can be efficiently learned in a self-supervised setting. 
The qualitative results in \cref{fig:qual_results} and the Supplement indicate that the 3D tasks were often unsolvable without our contributions. 

\subsection{Ablations on our Methods}
\label{sec:ablation}

\begin{table}[t]
\footnotesize
\caption{Ablation study on our proposed Loss ($\mathcal{L}_{\mathrm{Reslt}}$) and \ac{da} framework in a transfer learning setting (experiments congruent to \cref{table:main_results}). Performance improvements are associated with both our loss and \ac{da}. Both components combined lead to the best results.}
\centering
\begin{tabular}{c|c|c|c|c|c|c}
\toprule
\multicolumn{1}{l|}{\textbf{\begin{tabular}[c]{@{}c@{}}Exper-\\ iment\end{tabular}}} & \multicolumn{1}{c|}{$\mathbf{\mathcal{L}_{\mathrm{Reslt}}}$} & \multicolumn{1}{c|}{\textbf{DA}} & \multicolumn{1}{c|}{\textbf{\begin{tabular}[c]{@{}c@{}}Node\\ mAP\end{tabular}}} & \multicolumn{1}{c|}{\textbf{\begin{tabular}[c]{@{}c@{}}Node\\ mAR\end{tabular}}} & \multicolumn{1}{c|}{\textbf{\begin{tabular}[c]{@{}c@{}}Edge\\ mAP\end{tabular}}} & \multicolumn{1}{c}{\textbf{\begin{tabular}[c]{@{}c@{}}Edge\\ mAR\end{tabular}}} \\ \midrule
\multirow{3}{*}{\textbf{C}} 
 & \ding{55} & \ding{55} & 0.389 & 0.475 & 0.294 & 0.383 \\
 & \ding{55} & \ding{51} & 0.456 & 0.538 & 0.351 & 0.464\\
 & \ding{51} & \ding{51} & \textbf{0.491} & \textbf{0.571} & \textbf{0.366} & \textbf{0.489} \\ 
 \hline
 \multirow{3}{*}{\textbf{D}} 
 & \ding{55} & \ding{55} & 0.190 & 0.285 & 0.122 & 0.210 \\
 & \ding{55} & \ding{51} & 0.349 & 0.443 & 0.219 & 0.320 \\
 & \ding{51} & \ding{51} & \textbf{0.356} & \textbf{0.450} & \textbf{0.221} & \textbf{0.322} \\ \bottomrule
\end{tabular}
\label{table:main_ablations}

\end{table}

\begin{table}[t]
\footnotesize
\caption{Ablation study on our proposed Loss ($\mathcal{L}_{\mathrm{Reslt}}$) with and without pretraining (experiments congruent to \cref{table:main_results}). We find that our loss improves performance in an arbitrary image-to-graph learning setting. Here, the \ac{da} is employed in all experiments.}
\centering
\begin{tabular}{c|c|c|c|c|c|c}
\toprule
\multicolumn{1}{l|}{\textbf{\begin{tabular}[c]{@{}c@{}}Exper-\\ iment\end{tabular}}} & \multicolumn{1}{c|}{\textbf{\begin{tabular}[c]{@{}c@{}}Pre-\\ training\end{tabular}}} & \multicolumn{1}{c|}{\textbf{$\mathbf{\mathcal{L}_{\mathrm{Reslt}}}$}} & \multicolumn{1}{c|}{\textbf{\begin{tabular}[c]{@{}c@{}}Node\\ mAP\end{tabular}}} & \multicolumn{1}{c|}{\textbf{\begin{tabular}[c]{@{}c@{}}Node\\ mAR\end{tabular}}} & \multicolumn{1}{c|}{\textbf{\begin{tabular}[c]{@{}c@{}}Edge\\ mAP\end{tabular}}} & \multicolumn{1}{c}{\textbf{\begin{tabular}[c]{@{}c@{}}Edge\\ mAR\end{tabular}}} \\ \midrule
\multirow{4}{*}{\textbf{E}}
 & \ding{55} & \ding{55} & 0.231 & 0.308 & 0.249 & 0.329 \\
 & \ding{55} & \ding{51} & 0.410 & 0.464 & 0.468 & 0.512 \\
 & \ding{51} & \ding{55} & 0.424 & 0.488 & 0.449 & 0.510 \\
 & \ding{51} & \ding{51} & \textbf{0.483} & \textbf{0.535} & \textbf{0.523} & \textbf{0.566} \\ 
 \bottomrule
\end{tabular}
\label{table:loss_ablation}

\end{table}

In \cref{table:main_ablations} and \ref{table:loss_ablation}, we present ablations on the regularized edge sampling loss (\ref{sec:loss}) and \ac{da} framework (\ref{sec:domain_adaption}) for the 2D OCTA-500 \cite{li2020octa500} (\cref{table:main_ablations} C), the 3D MRI \cite{schneider2012tissue} (\cref{table:main_ablations} D), and the 3D brain vessel dataset \cite{todorov2020machine} (\cref{table:loss_ablation} E). 
Expectedly, we observe that our \ac{da} alone leads to compelling performance gains for the 3D setting (almost double the performance) and 2D setting across all metrics; see \cref{table:main_ablations}. This is expected since the domain shift between the source dataset of satellite images and our medical images is large. Note that our projection function is always employed for the 3D dataset since pretraining is otherwise impossible. We further observe that employing the projection function alone diminishes the performance because of the domain gap, which is only alleviated by our other contributions.

Next, we ablate on our loss. When applying the \ac{da}, adding our $\mathcal{L}_{\mathrm{Reslt}}$ loss further improves the performance (\cref{table:main_ablations}), indicating its strength in stabilizing and improving the loss landscape to train better networks. Additionally, we ablate our proposed $\mathcal{L}_{\mathrm{Reslt}}$ in an experiment with and without \ac{tl}. Importantly, our experiments show that $\mathcal{L}_{\mathrm{Reslt}}$ is a general contribution that improves image-to-graph transformation for \ac{tl} as well as for general network training (\cref{table:loss_ablation}). In a \ac{tl} setting, $\mathcal{L}_{\mathrm{Reslt}}$ is particularly useful as it reduces the data-specific hyperparameter search. Interestingly, our loss improves not only edge detection but also node detection metrics across our ablations. These improvements can be attributed to the transformer's cross-attention modules, which treat node and edge detection as joint prediction tasks instead of separate problems. Consequently, both metrics improve jointly. For further ablation studies, e.g., experiments without the domain adversarial networks or an alternative over-sampling of the parameter $r$, please refer to the Supplement. In conclusion, we note that each individual contribution enhances the overall performance of the graph prediction task.

\section{Discussion and Conclusion}

In this work, we propose a framework for cross-domain and cross-dimension transfer learning for image-to-graph transformers. At the core of this work are our strong empirical results, which show that our proposed inductive transfer learning method outperforms competing approaches across six benchmark datasets that contain 2D and 3D images by a margin. We achieve these results through our three methodological contributions, which we ablate individually. 
We conclude that transfer learning has the potential to substantially reduce data requirements for highly complex geometric deep learning tasks, such as transformer-based image-to-graph inference, see Supplement.
Our work shows that this holds especially when the targeted graph representations are defined by a similar physical principle or physical network. In the presented work, this shared principle is the transport of physical units (cars and blood) in a physical network. 

\paragraph{Limitations and future work.} 
Our problem setting is specific to image-to-graph tasks and our learning scenario in which we have some labeled data in both domains. Future work should investigate how our solution translates to different settings. Furthermore, we use a dimensionality-dependent feature extractor, which might limit generalizability to other dimensions. Future work should explore the development of strong dimension-invariant graph extractors to allow further generalization.

\vspace{0.2cm}
\small
\noindent\textbf{Acknowledgments.} AB is supported by the Stiftung der Deutschen Wirtschaft. MM is funded by the German Research Foundation under project 532139938. GK received support from the German Ministry of Education and Research and the Medical Informatics Initiative as part of the PrivateAIM Project and from the Bavarian Collaborative Research Project PRIPREKI of the Free State of Bavaria Funding Programme "Artificial Intelligence -- Data Science". This work was partially funded by the Konrad Zuse School of Excellence in Reliable AI (relAI).

{\small
\bibliographystyle{ieee_fullname}
\bibliography{main}

\begin{thebibliography}{10}\itemsep=-1pt

\bibitem{almasi2015graphextractionmedical}
Sepideh Almasi, Xiaoyin Xu, Ayal Ben-Zvi, Baptiste Lacoste, Chenghua Gu, and Eric~L Miller.
\newblock A novel method for identifying a graph-based representation of 3-d microvascular networks from fluorescence microscopy image stacks.
\newblock {\em Medical image analysis}, 20(1):208--223, 2015.

\bibitem{bastani2018roadtracer}
Favyen Bastani, Songtao He, Sofiane Abbar, Mohammad Alizadeh, Hari Balakrishnan, Sanjay Chawla, Sam Madden, and David DeWitt.
\newblock Roadtracer: Automatic extraction of road networks from aerial images.
\newblock In {\em Proceedings of the IEEE Conference on Computer Vision and Pattern Recognition}, pages 4720--4728, 2018.

\bibitem{batten2023imagetotree}
James Batten, Matthew Sinclair, Ben Glocker, and Michiel Schaap.
\newblock Image to tree with recursive prompting, 2023.

\bibitem{belli2019nodeprune}
Davide Belli and Thomas Kipf.
\newblock Image-conditioned graph generation for road network extraction, 2019.

\bibitem{ben2010theory}
Shai Ben-David, John Blitzer, Koby Crammer, Alex Kulesza, Fernando Pereira, and Jennifer~Wortman Vaughan.
\newblock A theory of learning from different domains.
\newblock {\em Machine learning}, 79:151--175, 2010.

\bibitem{biagioni2012topo}
James Biagioni and Jakob Eriksson.
\newblock Inferring road maps from global positioning system traces: Survey and comparative evaluation.
\newblock {\em Transportation Research Record}, 2291(1):61--71, 2012.

\bibitem{carion2020detr}
Nicolas Carion, Francisco Massa, Gabriel Synnaeve, Nicolas Usunier, Alexander Kirillov, and Sergey Zagoruyko.
\newblock End-to-end object detection with transformers.
\newblock In {\em Computer Vision--ECCV 2020: 16th European Conference, Glasgow, UK, August 23--28, 2020, Proceedings, Part I 16}, pages 213--229. Springer, 2020.

\bibitem{chen2021retinal}
Chunhui Chen, Joon~Huang Chuah, Raza Ali, and Yizhou Wang.
\newblock Retinal vessel segmentation using deep learning: a review.
\newblock {\em IEEE Access}, 9:111985--112004, 2021.

\bibitem{chen2021empirical}
X. Chen, S. Xie, and K. He.
\newblock An empirical study of training self-supervised vision transformers.
\newblock In {\em 2021 IEEE/CVF International Conference on Computer Vision (ICCV)}, pages 9620--9629, Los Alamitos, CA, USA, oct 2021. IEEE Computer Society.

\bibitem{chen2018domain}
Yuhua Chen, Wen Li, Christos Sakaridis, Dengxin Dai, and Luc Van~Gool.
\newblock Domain adaptive faster r-cnn for object detection in the wild.
\newblock In {\em Proceedings of the IEEE conference on computer vision and pattern recognition}, pages 3339--3348, 2018.

\bibitem{chen2022road}
Ziyi Chen, Liai Deng, Yuhua Luo, Dilong Li, Jos{\'e}~Marcato Junior, Wesley~Nunes Gon{\c{c}}alves, Abdul Awal~Md Nurunnabi, Jonathan Li, Cheng Wang, and Deren Li.
\newblock Road extraction in remote sensing data: A survey.
\newblock {\em International Journal of Applied Earth Observation and Geoinformation}, 112:102833, 2022.

\bibitem{Dai2021upDETR}
Zhigang Dai, Bolun Cai, Yugeng Lin, and Junying Chen.
\newblock Up-detr: Unsupervised pre-training for object detection with transformers.
\newblock In {\em Proceedings of the IEEE/CVF Conference on Computer Vision and Pattern Recognition (CVPR)}, pages 1601--1610, June 2021.

\bibitem{dosovitskiy2020vit}
Alexey Dosovitskiy, Lucas Beyer, Alexander Kolesnikov, Dirk Weissenborn, Xiaohua Zhai, Thomas Unterthiner, Mostafa Dehghani, Matthias Minderer, Georg Heigold, Sylvain Gelly, et~al.
\newblock An image is worth 16x16 words: Transformers for image recognition at scale.
\newblock {\em arXiv preprint arXiv:2010.11929}, 2020.

\bibitem{drees2021voreen}
Dominik Drees, Aaron Scherzinger, Ren{\'e} H{\"a}gerling, Friedemann Kiefer, and Xiaoyi Jiang.
\newblock Scalable robust graph and feature extraction for arbitrary vessel networks in large volumetric datasets.
\newblock {\em BMC bioinformatics}, 22(1):1--28, 2021.

\bibitem{ermagun2018spatiotemporal}
Alireza Ermagun and David Levinson.
\newblock Spatiotemporal traffic forecasting: review and proposed directions.
\newblock {\em Transport Reviews}, 38(6):786--814, 2018.

\bibitem{erturk2012three}
Ali Ert{\"u}rk, Klaus Becker, Nina J{\"a}hrling, Christoph~P Mauch, Caroline~D Hojer, Jackson~G Egen, Farida Hellal, Frank Bradke, Morgan Sheng, and Hans-Ulrich Dodt.
\newblock Three-dimensional imaging of solvent-cleared organs using 3disco.
\newblock {\em Nature protocols}, 7(11):1983--1995, 2012.

\bibitem{ganin2015unsupervised}
Yaroslav Ganin and Victor Lempitsky.
\newblock Unsupervised domain adaptation by backpropagation.
\newblock In {\em International conference on machine learning}, pages 1180--1189. PMLR, 2015.

\bibitem{glorot2010understanding}
Xavier Glorot and Yoshua Bengio.
\newblock Understanding the difficulty of training deep feedforward neural networks.
\newblock In {\em Proceedings of the thirteenth international conference on artificial intelligence and statistics}, pages 249--256. JMLR Workshop and Conference Proceedings, 2010.

\bibitem{googleapi}
Google.
\newblock Google static maps api.

\bibitem{haklay2008osm}
Mordechai Haklay and Patrick Weber.
\newblock Openstreetmap: User-generated street maps.
\newblock {\em IEEE Pervasive Computing}, 7(4):12--18, 2008.

\bibitem{hammami2020cycle}
Maryam Hammami, Denis Friboulet, and Razmig K{\'e}chichian.
\newblock Cycle gan-based data augmentation for multi-organ detection in ct images via yolo.
\newblock In {\em 2020 IEEE international conference on image processing (ICIP)}, pages 390--393. IEEE, 2020.

\bibitem{he2020sat2graph}
Songtao He, Favyen Bastani, Satvat Jagwani, Mohammad Alizadeh, Hari Balakrishnan, Sanjay Chawla, Mohamed~M Elshrif, Samuel Madden, and Mohammad~Amin Sadeghi.
\newblock Sat2graph: Road graph extraction through graph-tensor encoding.
\newblock In {\em European Conference on Computer Vision}, pages 51--67. Springer, 2020.

\bibitem{huang2018auggan}
Sheng-Wei Huang, Che-Tsung Lin, Shu-Ping Chen, Yen-Yi Wu, Po-Hao Hsu, and Shang-Hong Lai.
\newblock Auggan: Cross domain adaptation with gan-based data augmentation.
\newblock In {\em Proceedings of the European Conference on Computer Vision (ECCV)}, pages 718--731, 2018.

\bibitem{kornblith2019similarity}
Simon Kornblith, Mohammad Norouzi, Honglak Lee, and Geoffrey Hinton.
\newblock Similarity of neural network representations revisited.
\newblock In {\em International conference on machine learning}, pages 3519--3529. PMLR, 2019.

\bibitem{kreitner2023detailed}
Linus Kreitner, Johannes~C Paetzold, Nikolaus Rauch, Chen Chen, Ahmed~M Hagag, Alaa~E Fayed, Sobha Sivaprasad, Sebastian Rausch, Julian Weichsel, Bjoern~H Menze, et~al.
\newblock Detailed retinal vessel segmentation without human annotations using simulated optical coherence tomography angiographs.
\newblock {\em arXiv preprint arXiv:2306.10941}, 2023.

\bibitem{li2020octa500}
Mingchao Li, Yuhan Zhang, Zexuan Ji, Keren Xie, Songtao Yuan, Qinghuai Liu, and Qiang Chen.
\newblock Ipn-v2 and octa-500: Methodology and dataset for retinal image segmentation.
\newblock {\em arXiv preprint arXiv:2012.07261}, 2020.

\bibitem{li2022graph}
Michelle~M Li, Kexin Huang, and Marinka Zitnik.
\newblock Graph representation learning in biomedicine and healthcare.
\newblock {\em Nature Biomedical Engineering}, pages 1--17, 2022.

\bibitem{Li2021transferViT}
Yanghao Li, Saining Xie, Xinlei Chen, Piotr Doll{\'a}r, Kaiming He, and Ross~B. Girshick.
\newblock Benchmarking detection transfer learning with vision transformers.
\newblock {\em ArXiv}, abs/2111.11429, 2021.

\bibitem{liu2021pixeltopoint}
Yueh-Cheng Liu, Yu-Kai Huang, Hung-Yueh Chiang, Hung-Ting Su, Zhe-Yu Liu, Chin-Tang Chen, Ching-Yu Tseng, and Winston~H. Hsu.
\newblock Learning from 2d: Contrastive pixel-to-point knowledge transfer for 3d pretraining, 2021.

\bibitem{long2018conditional}
Mingsheng Long, Zhangjie Cao, Jianmin Wang, and Michael~I Jordan.
\newblock Conditional adversarial domain adaptation.
\newblock {\em Advances in neural information processing systems}, 31, 2018.

\bibitem{lucic2018gans}
Mario Lucic, Karol Kurach, Marcin Michalski, Sylvain Gelly, and Olivier Bousquet.
\newblock Are gans created equal? a large-scale study.
\newblock {\em Advances in neural information processing systems}, 31, 2018.

\bibitem{ma2021rose}
Yuhui Ma, Huaying Hao, Jianyang Xie, Huazhu Fu, Jiong Zhang, Jianlong Yang, Zhen Wang, Jiang Liu, Yalin Zheng, and Yitian Zhao.
\newblock Rose: a retinal oct-angiography vessel segmentation dataset and new model.
\newblock {\em IEEE Transactions on Medical Imaging}, 40(3):928--939, 2021.

\bibitem{ma2023revisiting}
Yan Ma, Weicong Liang, Yiduo Hao, Bohan Chen, Xiangyu Yue, Chao Zhang, and Yuhui Yuan.
\newblock Revisiting detr pre-training for object detection.
\newblock {\em arXiv preprint arXiv:2308.01300}, 2023.

\bibitem{martin2020graph}
Henry Martin, Dominik Bucher, Ye Hong, Ren{\'e} Buffat, Christian Rupprecht, and Martin Raubal.
\newblock Graph-resnets for short-term traffic forecasts in almost unknown cities.
\newblock In {\em NeurIPS 2019 Competition and Demonstration Track}, pages 153--163. PMLR, 2020.

\bibitem{menten2022physiology}
Martin~J Menten, Johannes~C Paetzold, Alina Dima, Bjoern~H Menze, Benjamin Knier, and Daniel Rueckert.
\newblock Physiology-based simulation of the retinal vasculature enables annotation-free segmentation of oct angiographs.
\newblock In {\em International Conference on Medical Image Computing and Computer-Assisted Intervention (MICCAI)}, pages 330--340. Springer, 2022.

\bibitem{padilla2020detMetrics}
Rafael Padilla, Sergio~L. Netto, and Eduardo A.~B. da Silva.
\newblock A survey on performance metrics for object-detection algorithms.
\newblock In {\em 2020 International Conference on Systems, Signals and Image Processing (IWSSIP)}, pages 237--242, 2020.

\bibitem{paetzold2021whole}
Johannes~C Paetzold, Julian McGinnis, Suprosanna Shit, Ivan Ezhov, Paul B{\"u}schl, Chinmay Prabhakar, Anjany Sekuboyina, Mihail Todorov, Georgios Kaissis, Ali Ert{\"u}rk, et~al.
\newblock Whole brain vessel graphs: A dataset and benchmark for graph learning and neuroscience.
\newblock In {\em Thirty-fifth Conference on Neural Information Processing Systems Datasets and Benchmarks Track (Round 2)}, 2021.

\bibitem{pan2009survey}
Sinno Pan and Qiang Yang.
\newblock A survey on transfer learning.
\newblock {\em IEEE Transactions on knowledge, data engineering}, 2009.

\bibitem{prabhakar2023vesselformer}
Chinmay Prabhakar, Suprosanna Shit, Johannes~C Paetzold, Ivan Ezhov, Rajat Koner, Hongwei Li, Florian~Sebastian Kofler, et~al.
\newblock Vesselformer: Towards complete 3d vessel graph generation from images.
\newblock In {\em Medical Imaging with Deep Learning}, 2023.

\bibitem{schneider2012tissue}
Matthias Schneider et~al.
\newblock Tissue metabolism driven arterial tree generation.
\newblock {\em Med Image Anal.}, 16(7):1397--1414, 2012.

\bibitem{shen2022simcrosstrans}
Xiaoke Shen and Ioannis Stamos.
\newblock simcrosstrans: A simple cross-modality transfer learning for object detection with convnets or vision transformers, 2022.

\bibitem{shit2022relationformer}
Suprosanna Shit, Rajat Koner, Bastian Wittmann, Johannes Paetzold, Ivan Ezhov, Hongwei Li, Jiazhen Pan, Sahand Sharifzadeh, Georgios Kaissis, Volker Tresp, et~al.
\newblock Relationformer: A unified framework for image-to-graph generation.
\newblock In {\em ECCV 2022: 17th European Conference on Computer Vision, October 2022}, 2022.

\bibitem{shit2022graflow}
Suprosanna Shit, Chinmay Prabhakar, Johannes~C Paetzold, Martin~J Menten, Bastian Wittmann, Ivan Ezhov, et~al.
\newblock Graflow: Neural blood flow solver for vascular graph.
\newblock In {\em Geometric Deep Learning in Medical Image Analysis (Extended abstracts)}, 2022.

\bibitem{sobisch2022automated}
Jannik Sobisch, {\v{Z}}iga Bizjak, Aichi Chien, and {\v{Z}}iga {\v{S}}piclin.
\newblock Automated intracranial vessel labeling with learning boosted by vessel connectivity, radii and spatial context.
\newblock In {\em Geometric Deep Learning in Medical Image Analysis}, pages 34--44. PMLR, 2022.

\bibitem{song2019graph_altzheimer}
Tzu-An Song, Samadrita~Roy Chowdhury, Fan Yang, Heidi Jacobs, Georges El~Fakhri, Quanzheng Li, Keith Johnson, and Joyita Dutta.
\newblock Graph convolutional neural networks for alzheimer’s disease classification.
\newblock In {\em 2019 IEEE 16th international symposium on biomedical imaging (ISBI 2019)}, pages 414--417. IEEE, 2019.

\bibitem{tetteh2018deepvesselnet}
Giles Tetteh et~al.
\newblock Deepvesselnet: Vessel segmentation, centerline prediction, and bifurcation detection in 3-d angiographic volumes.
\newblock {\em arXiv preprint arXiv:1803.09340}, 2018.

\bibitem{todorov2020machine}
Mihail~Ivilinov Todorov*, Johannes Paetzold*, Oliver Schoppe, Giles Tetteh, Suprosanna Shit, Velizar Efremov, Katalin Todorov-V{\"o}lgyi, Marco D{\"u}ring, Martin Dichgans, Marie Piraud, et~al.
\newblock Machine learning analysis of whole mouse brain vasculature.
\newblock {\em Nature Methods}, 17(4):442--449, 2020.

\bibitem{tzeng2017adversarial}
Eric Tzeng, Judy Hoffman, Kate Saenko, and Trevor Darrell.
\newblock Adversarial discriminative domain adaptation.
\newblock In {\em Proceedings of the IEEE conference on computer vision and pattern recognition}, pages 7167--7176, 2017.

\bibitem{wang2020classes}
Haoran Wang, Tong Shen, Wei Zhang, Ling-Yu Duan, and Tao Mei.
\newblock Classes matter: A fine-grained adversarial approach to cross-domain semantic segmentation.
\newblock In {\em European conference on computer vision}, pages 642--659. Springer, 2020.

\bibitem{wang2018deep}
Mei Wang and Weihong Deng.
\newblock Deep visual domain adaptation: A survey.
\newblock {\em Neurocomputing}, 312:135--153, 2018.

\bibitem{wang2016roadreview}
Weixing Wang, Nan Yang, Yi Zhang, Fengping Wang, Ting Cao, and Patrik Eklund.
\newblock A review of road extraction from remote sensing images.
\newblock {\em Journal of traffic and transportation engineering (english edition)}, 3(3):271--282, 2016.

\bibitem{wang2022vision2dto3d}
Yi Wang, Zhiwen Fan, Tianlong Chen, Hehe Fan, and Zhangyang Wang.
\newblock Can we solve 3d vision tasks starting from a 2d vision transformer?, 2022.

\bibitem{xie2021unified2d3d}
Yutong Xie, Jianpeng Zhang, Yong Xia, and Qi Wu.
\newblock Unified 2d and 3d pre-training for medical image classification and segmentation.
\newblock {\em arXiv preprint arXiv:2112.09356}, 2021.

\bibitem{xu2022rngdet++}
Zhenhua Xu, Yuxuan Liu, Yuxiang Sun, Ming Liu, and Lujia Wang.
\newblock Rngdet++: Road network graph detection by transformer with instance segmentation and multi-scale features enhancement.
\newblock {\em arXiv preprint arXiv:2209.10150}, 2022.

\bibitem{zhou2019histology3}
Yanning Zhou, Simon Graham, Navid Alemi~Koohbanani, Muhammad Shaban, Pheng-Ann Heng, and Nasir Rajpoot.
\newblock Cgc-net: Cell graph convolutional network for grading of colorectal cancer histology images.
\newblock In {\em Proceedings of the IEEE/CVF international conference on computer vision workshops}, 2019.

\bibitem{zhu2017unpaired}
Jun-Yan Zhu, Taesung Park, Phillip Isola, and Alexei~A Efros.
\newblock Unpaired image-to-image translation using cycle-consistent adversarial networks.
\newblock In {\em Proceedings of the IEEE international conference on computer vision}, pages 2223--2232, 2017.

\end{thebibliography}
}

\setcounter{page}{1}
{
\centering
\Large
\noindent\\
\textbf{Supplementary Material} \\
\vspace{0.5em}
}

\centerline{\large Cross-domain and Cross-dimension Learning}
\centerline{\large for Image-to-Graph Transformers}

\section{Additional ablation studies}
Additionally to our main experiments, we present ablation studies on further aspects of our proposed framework. These ablation studies give a deeper understanding of the components' dynamics and guide future reimplementations and adaptions. 

\subsection{Generalizability}
\begin{table*}[ht]
    \centering
    \small
    \caption{Ablation study on the pretraining dataset. As the domain gap between the source and target domain decreases, the downstream performance increases. We show that our method is generalizable across different pretraining datasets.}
    \begin{tabular}{ l | l c c c c c }
        \toprule
        \begin{tabular}[c]{@{}l@{}} \textbf{Fine Tuning} \\ \textbf{Training Set}\end{tabular} 
        & \begin{tabular}[c]{@{}l@{}} \textbf{(Pre-)Training} \\ \textbf{Strategy}\end{tabular} 
        & \begin{tabular}[c]{@{}l@{}} \textbf{Node-}     \\ \textbf{mAP}$\uparrow$\end{tabular} 
        & \begin{tabular}[c]{@{}l@{}} \textbf{Node-}     \\ \textbf{mAR}$\uparrow$\end{tabular} 
        & \begin{tabular}[c]{@{}l@{}} \textbf{Edge-}     \\ \textbf{mAP}$\uparrow$\end{tabular} 
        & \begin{tabular}[c]{@{}l@{}} \textbf{Edge-}     \\ \textbf{mAR}$\uparrow$\end{tabular} 
        & \textbf{SMD $\downarrow$} \\
    
        \midrule
        \multirow{5}{*}{\begin{tabular}[c]{@{}l@{}} Microscopic \\ images \cite{todorov2020machine}\end{tabular}}
        &  No Pretraining \cite{glorot2010understanding} & 0.231 & 0.308 & 0.249 & 0.329 & 0.017\\
        &  Self-supervised \cite{chen2021empirical}      & 0.344 & 0.404 & 0.363 & 0.425 & 0.017\\
        &  Supervised                                    & $\star$  & $\star$  & $\star$  & $\star$  & $\star$ \\
        &  Ours, pretr. on cities                        & 0.483 & 0.535 & 0.523 & 0.566 & 0.017\\
        &  \textbf{Ours, pretr. on OCTA}                 & \textbf{0.548} & \textbf{0.583} & \textbf{0.588} & \textbf{0.615} & \textbf{0.016}\\
        \bottomrule
    \end{tabular}
    \label{table:generalizability_ablation}
\end{table*}

In Table \ref{table:generalizability_ablation}, we present the results of our experiment E) (see Table \ref{table:main_results}) with one additional configuration. In this configuration, we pretrain the model on the synthetic OCTA dataset \cite{menten2022physiology} instead of the U.S. cities dataset \cite{he2020sat2graph}. This pretraining strategy outperforms all baselines and increases the performance compared to our main experimental setting (see Section \ref{sec:results}). We attribute this improvement to the smaller domain gap between the new source domain (i.e., retinal blood vessels) and the target domain (i.e., a mouse's cerebrovasculature). These results show how our method generalizes seamlessly to new domains. Furthermore, they substantiate the rationale behind our experimental design: by showcasing the utility of our method in a challenging setting, focused on the most intricate transfer learning scenarios, we establish its effectiveness in more straightforward transfer learning situations (as presented in Table \ref{table:generalizability_ablation} as well.

\subsection{Regularized edge sampling loss}
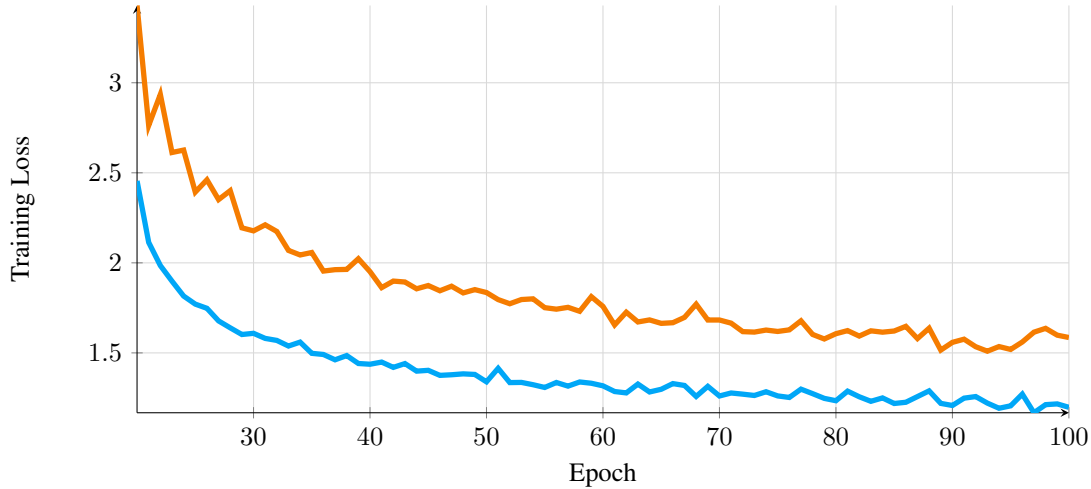
\begin{figure*}[ht]
\centering
\begin{tikzpicture}
    \begin{axis}[
                 grid=both,
                 grid style={solid,gray!30!white},
                 axis lines=middle,
                 xlabel={Epoch},
                 ylabel={Training Loss},
                 x label style={at={(axis description cs:0.5,-0.1)},anchor=north},
                 y label style={at={(axis description cs:-0.1,.5)},rotate=90,anchor=south},
                 width=0.8\linewidth, height=7cm]
        \addplot[line width=2pt, color=tb_color_1] table [x=Step, y=Value, col sep=comma] {figs/train_loss_curve/suppl_normal-loss_curve.csv};
        \addplot[line width=2pt, color=tb_color_2] table [x=Step, y=Value, col sep=comma] {figs/train_loss_curve/suppl_our-loss_curve.csv};
    \end{axis}
  \end{tikzpicture}
  \caption{Training loss curves. The orange line depicts the training loss without our regularized edge sampling loss $\mathcal{L}_{Resln}$ and the blue line with $\mathcal{L}_{Resln}$, respectively. $\mathcal{L}_{Resln}$ shows faster convergence from the beginning on.}
\label{fig:training_curves}
\end{figure*}

\begin{table}[ht]
    \small
    \centering
    \caption{Ablation study on the loss ratio $r$ for $\mathcal{L}_{Resln}$ as described in Section \ref{sec:loss} (experiments congruent to Table \ref{table:main_results}). We observe that $\mathcal{L}_{Resln}$ is stable across varying loss ratios and does not require sensitive hyperparameter tuning.}
    \begin{tabular}{c|ccccc}
    \toprule
    \multicolumn{1}{l|}{\textbf{\begin{tabular}[c]{@{}c@{}}Experi-\\ ment\end{tabular}}} & $\boldsymbol{r}$ & \multicolumn{1}{c}{\textbf{\begin{tabular}[c]{@{}c@{}}Node\\ mAP\end{tabular}}} & \multicolumn{1}{c}{\textbf{\begin{tabular}[c]{@{}c@{}}Node\\ mAR\end{tabular}}} & \multicolumn{1}{c}{\textbf{\begin{tabular}[c]{@{}c@{}}Edge\\ mAP\end{tabular}}} & \multicolumn{1}{c}{\textbf{\begin{tabular}[c]{@{}c@{}}Edge\\ mAR\end{tabular}}} \\ \midrule
    \multirow{7}{*}{\textbf{D}} 
     & 0.05 & 0.3539 & 0.4446 & 0.2166 & 0.3102 \\
     & 0.1 & \textbf{0.3564} & \textbf{0.4498} & 0.2209 & 0.3218 \\
     & 0.15 & 0.3470 & 0.4380 & 0.2164 & 0.3122 \\
     & 0.2 & 0.3532 & 0.4449 & 0.2203 & 0.3193 \\
     & 0.3 & 0.3451 & 0.4351 & 0.2183 & 0.3153 \\
     & 0.5 & 0.3470 & 0.4407 & 0.2231 & 0.3242 \\
     & 0.8 & 0.3462 & 0.4394 & \textbf{0.2253} & \textbf{0.3288} \\
     \bottomrule
    \end{tabular}
    \label{table:loss_ratio_ablation}
\end{table}

Next, we conduct an ablation study on the effect of the regularized edge sampling loss. As explained in Section \ref{sec:results}, the new loss stabilizes training and increases convergence speed. This effect is shown in Figure \ref{fig:training_curves}, where the training loss decreases faster from the beginning on and convergences towards a lower level compared to the baseline loss formulation. This effect can be observed across all datasets and training strategies. Also, we experiment with different foreground-to-background-edge ratios $r$ (see Section \ref{sec:loss}). Table \ref{table:loss_ratio_ablation} shows that the performance stays stable across a large range of $r$-values. These results underline our hypothesis from Section \ref{sec:loss} that $\mathcal{L}_{Resln}$ reduces the hyperparameter space because it does not require careful optimization.

\begin{table*}[ht]
    \small
    \centering
    \caption{Ablation study on different edge sampling strategies. }
    \begin{tabular}{l|cc|cc}
    \toprule
    \multirow{2}{*}{\textbf{Strategy}} & \multicolumn{2}{c|}{\textbf{Experiment A}} & \multicolumn{2}{c}{\textbf{Experiment E}} \\
    & \textbf{node-mAP} & \textbf{edge-mAP} & \textbf{node-mAP} & \textbf{edge-mAP} \\
    \midrule
    \textbf{subsampling} & 0.172 & 0.125 & 0.237 & 0.277\\
    \textbf{varying-$r$} & 0.156 & 0.115 & 0.218 & 0.134\\
    \textbf{oversampling (ours)} & 0.173 & 0.129 & 0.267 & 0.323\\
    \bottomrule
    \end{tabular}
    \label{table:loss_sampling_strategy}
\end{table*}

Furthermore, we study the effect of different edge-sampling strategies on our loss formulation in Table \ref{table:loss_sampling_strategy}. Specifically, we compare our fixed-ratio upsampling strategy with a varying-$r$ upsampling (i.e., for each batch, we randomly choose $r$ with a uniform distribution in $(,1]$), and a fixed-ratio subsampling strategy. The decreased performance with a varying-$r$ upsampling strategy shows that a fixed $r$ is important for our loss formulation. We further find that subsampling is a valid alternative in scenarios where data is extremely scarce (e.g., Experiment A) but performs worse when more data is available (e.g., Experiment E). Notably, Shit et al. \cite{shit2022relationformer} proposed a one-sided subsampling strategy, i.e., subsampling only the background edges if the ratio is \textbf{above} a certain ratio. This strategy is problematic when the target dataset contains dense graphs, in which our loss formulation upsamples the background edges (see Table \ref{table:edge_stats} for dataset statistics). Furthermore, the official relationformer repository does use a dynamic subsampling strategy but selects background edges up to an absolute threshold $m$, which introduces strong hyperparameter sensitivity. Table \ref{table:edge_stats} shows that up- or sub-sampling only one edge type (e.g., the background edges) would not be sufficient.

\begin{table}[ht]
\centering
\scriptsize
\caption{Edge statistics for the used datasets. The varying ratios between active and background edges underline the utility of our dynamic loss formulation. Different datasets require upsampling for active and background edges.}
\label{table:edge_stats}
\begin{tabular}{l|cccc}
\toprule
\textbf{Dataset}        & \textbf{\begin{tabular}[c]{@{}c@{}}Avg.\\ Edges\end{tabular}} & \textbf{\begin{tabular}[c]{@{}c@{}}Avg. edge\\ ratio\end{tabular}} & \textbf{\begin{tabular}[c]{@{}c@{}}Upsampling\\ background\end{tabular}} & \textbf{\begin{tabular}[c]{@{}c@{}}Upsampling\\ active\end{tabular}} \\
\midrule
\textbf{20 U.S. Cities} & 6.37                & 0.53                     & 92.5\%                         & 7.5\%                      \\
\textbf{Agadez}         & 5.05                & 0.75                     & 96.2\%                         & 3.8\%                      \\
\textbf{Munich}         & 4.69                & 0.77                     & 97.1\%                         & 2.9\%                      \\
\textbf{Synth. OCTA}    & 32.57               & 0.05                     & 1.4\%                          & 98.6\%                     \\
\textbf{OCTA-500}       & 11.07               & 0.18                     & 47.5\%                         & 52.5\%                     \\
\textbf{Synth. MRI}     & 22.37               & 0.07                     & 0.3\%                          & 99.7\%                     \\
\textbf{Microscopy}     & 33.31               & 0.05                     & 0.3\%                          & 99.7\%                    \\
\bottomrule
\end{tabular}
\end{table}

\subsection{Domain adaptation framework}
\begin{table*}[ht]
\centering
\small
\caption{Ablation study on our domain adaptation framework in a transfer learning setting (experiments congruent to Table \ref{table:main_results}). $\mathcal{L}_{img}$, $\mathcal{L}_{graph}$, and $\mathcal{L}_{cst}$ refer to the optimization terms from Section \ref{sec:domain_adaption}. We find that performance improvements are associated with all adaptation components. Using the complete optimization term as presented in Section \ref{sec:complete_loss_formulation} yields the best results. }
\begin{tabular}{c|ccc|cccc}
\toprule
\multicolumn{1}{l|}{\textbf{\begin{tabular}[c]{@{}c@{}}Exper-\\ iment\end{tabular}}} & \multicolumn{1}{c}{$\mathcal{L}_{img}$} & \multicolumn{1}{c}{$\mathcal{L}_{graph}$} & \multicolumn{1}{c|}{$\mathcal{L}_{cst}$} & \multicolumn{1}{c}{\textbf{\begin{tabular}[c]{@{}c@{}}Node\\ mAP\end{tabular}}} & \multicolumn{1}{c}{\textbf{\begin{tabular}[c]{@{}c@{}}Node\\ mAR\end{tabular}}} & \multicolumn{1}{c}{\textbf{\begin{tabular}[c]{@{}c@{}}Edge\\ mAP\end{tabular}}} & \multicolumn{1}{c}{\textbf{\begin{tabular}[c]{@{}c@{}}Edge\\ mAR\end{tabular}}} \\ \midrule
\multirow{5}{*}{\textbf{C}} 
 & \ding{55} & \ding{55}& \ding{55} & 0.3423 & 0.4341 & 0.2581 & 0.3414 \\
 & \ding{51} & \ding{55}& \ding{55} & 0.4286 & 0.5073 & 0.3293 & 0.4264 \\
 & \ding{55} & \ding{51}& \ding{55} & 0.3264 & 0.4136 & 0.2355 & 0.3117 \\
 & \ding{51} & \ding{51}& \ding{55} & 0.4071 & 0.4980 & 0.2685 & 0.3831\\
 & \ding{51} & \ding{51}& \ding{51} & \textbf{0.4909} & \textbf{0.5712} & \textbf{0.3656} & \textbf{0.4887} \\ 
 \bottomrule
\end{tabular}
\label{table:adversarial_ablation}
\end{table*}

Table \ref{table:adversarial_ablation} shows an ablation study of our domain adaptation framework's components. $\mathcal{L}_{img}$, $\mathcal{L}_{graph}$, and $\mathcal{L}_{cst}$ refer to the optimization terms from Section \ref{sec:domain_adaption}. Using the image-level alignment alone already yields a performance increase of around 30 \% compared to not using our framework at all. We attribute this observation to the large image-level differences between the source and target domain, which hinders knowledge transfer in the feature extractor if an adversarial does not mitigate it. The graph-level adversarial slightly decreases the performance when being applied without consistency regularization (i.e., $\mathcal{L}_{cst}$). This decrease is likely caused by the abstraction level of the transformer's tokenized graph representation. Without any further guidance (e.g., by the image-level domain classifier through consistency regularization), the graph-level classifier does not provide a precise gradient toward a domain-invariant representation. Combining all three components yields the best results, supporting our hypothesis that the graph-level adversarial needs regularization by the image-level adversarial.

\begin{table}[ht]
\centering
\scriptsize
\caption{The performance of our contributions with and without our DA framework in the additional experiment from Table \ref{table:generalizability_ablation}. We show that smaller domain gaps (here, from OCTA to microscopy images) can be bridged without our DA framework even with dimension shift.}
\begin{tabular}{l|cccc}
\toprule
\textbf{Method}         & \textbf{node-mAP} & \textbf{node-mAR} & \textbf{edge-mAP} & \textbf{edge-mAR} \\ \midrule
\textbf{No Pretraining} & 0.231             & 0.308             & 0.249             & 0.329             \\
\textbf{No DA}          & 0.508             & 0.549             & 0.551             & 0.584             \\
\textbf{Ours}           & 0.548             & 0.583             & 0.588             & 0.615             \\
\bottomrule
\end{tabular}
\label{table:no_DA}
\end{table} 

Furthermore, we study the impact of our projection function and loss formulation without applying our domain adaptation framework. Table \ref{table:no_DA} shows that our other contributions alone enable transfer learning across dimensions. This enables transfer learning without access to the target domain during pretraining. However, even in these cases, our DA framework yields the best performance. Table \ref{table:main_ablations} shows a similar trend in cases without dimension shift.

\subsection{Adversarial learning coefficient}
\begin{table}[ht]
    \small
    \centering
    \caption{Ablation study on the domain adversarial learning coefficient $\alpha$ (experiments congruent to Table \ref{table:main_results}). $\alpha$ must be optimized such that the adversarial loss balances with the graph extraction loss. An $\alpha$ value of 0 is equivalent to not using the domain adaptation framework.}
    \begin{tabular}{c|ccccc}
    \toprule
    \multicolumn{1}{l|}{\textbf{\begin{tabular}[c]{@{}c@{}}Experi-\\ ment\end{tabular}}} & $\boldsymbol{\alpha}$ & \multicolumn{1}{c}{\textbf{\begin{tabular}[c]{@{}c@{}}Node\\ mAP\end{tabular}}} & \multicolumn{1}{c}{\textbf{\begin{tabular}[c]{@{}c@{}}Node\\ mAR\end{tabular}}} & \multicolumn{1}{c}{\textbf{\begin{tabular}[c]{@{}c@{}}Edge\\ mAP\end{tabular}}} & \multicolumn{1}{c}{\textbf{\begin{tabular}[c]{@{}c@{}}Edge\\ mAR\end{tabular}}} \\ \midrule
    \multirow{8}{*}{\textbf{C}} 
     & 0.0 & 0.3884 & 0.4755 & 0.2947 & 0.3767 \\
     & 0.1 & 0.3123 & 0.4076 & 0.2258 & 0.3044 \\
     & 0.3 & 0.3381 & 0.4287 & 0.2439 & 0.3240 \\
     & 0.5 & 0.4563 & 0.5395 & 0.3614 & 0.4618 \\ 
     & 0.8 & 0.4623 & 0.5458 & 0.3464 & 0.4687 \\
     & 1.0 & \textbf{0.4909} & \textbf{0.5712} & \textbf{0.3656} & \textbf{0.4887} \\
     & 1.5 & 0.3914 & 0.4726 & 0.2854 & 0.3897 \\
     & 2.0 & 0.0530 & 0.1719 & 0.0214 & 0.0451 \\
     \bottomrule
    \end{tabular}
    \label{table:alpha_ablation}
\end{table}

\begin{figure}[ht]
\centering
\begin{tikzpicture}
    \begin{axis}[grid=both,
                 grid style={solid,gray!30!white},
                 xlabel={Pretraining epoch},
                 ylabel={cka-similarity \cite{kornblith2019similarity}},
                 x label style={at={(axis description cs:0.5,0.0)},anchor=north},
                 y label style={at={(axis description cs:0.05,.5)},anchor=south},
                 legend style={at={(0.95,0.75)}},
                 ]
        \addplot[line width=2pt, color=tb_color_2] table [x=Step, y=Value, col sep=comma] {figs/cka_similarity_plot/cka_full5_real-eye_a3e-1_10.csv};
        \addlegendentry{$\alpha=0.3$}
        \addplot[line width=2pt, color=tb_color_1] table [x=Step, y=Value, col sep=comma] {figs/cka_similarity_plot/cka_full5_real-eye_a1_no-upsampling_10.csv};
        \addlegendentry{$\alpha=1.0$}
        \addplot[line width=2pt, color=tb_color_3] table [x=Step, y=Value, col sep=comma] {figs/cka_similarity_plot/cka_full5_real-eye_a15e-1_10.csv};
        \addlegendentry{$\alpha=1.5$}
    \end{axis}
  \end{tikzpicture}
  \caption{cka-similarity \cite{kornblith2019similarity} (y-axis) between the feature representations of source and target domain during pretraining. $alpha$ must be sufficiently large such that the similarity increases during training. From a certain threshold on, the similarity does not increase further. We associate a high similarity between both domains with the model learning domain-invariant features.}
\label{fig:suppl_alpha_curve}
\end{figure}
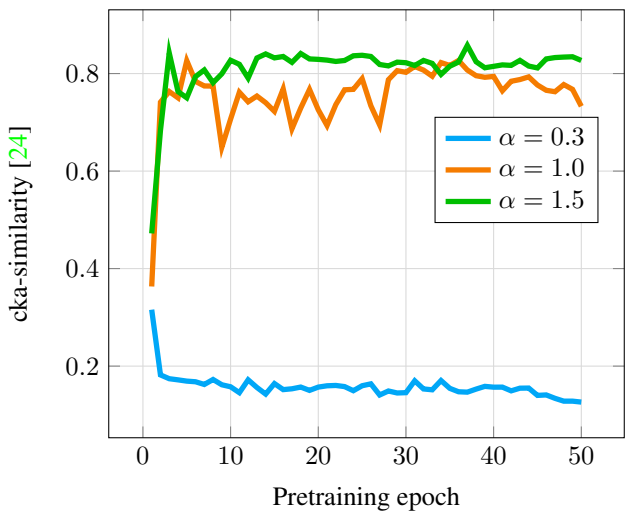

In Table \ref{table:alpha_ablation}, we ablate on the domain adversarial learning coefficient $\alpha$. $\alpha$ is the factor with which the gradient in the \ac{grl} is multiplied before passing it to the respective model component, i.e., the feature extractor for the image-level adversarial and the encoder-decoder for the graph-level adversarial (see Section \ref{sec:domain_adaption}). We use the $\alpha$ schedule proposed by Chen et al. \cite{chen2018domain}, which increases $\alpha$ during the training until reaching a fixed maximum. Table \ref{table:alpha_ablation} shows that the right choice of $\alpha$ is crucial and that a sub-optimal value can decrease downstream performance. We attribute this observation to the model's tradeoff between learning to produce domain-invariant features (i.e., domain confusion) and task learning (i.e., graph extraction). If $\alpha$ is too large, the adversarial loss dominates the task loss, and the network does not learn how to produce meaningful features. If it is too small, the domain gap between the source and target domain stays too large, and knowledge transfer is impeded. Figure \ref{fig:suppl_alpha_curve} shows how a small $\alpha$ (e.g., $\alpha=0.3$) is not sufficient to learn domain-invariant features while an $\alpha$-value that is too large does not increase domain confusion but obstructs learning the core task. Note that the specific $\alpha$ value must be optimized for the used datasets and is not domain-invariant.

\subsection{Target dataset size}
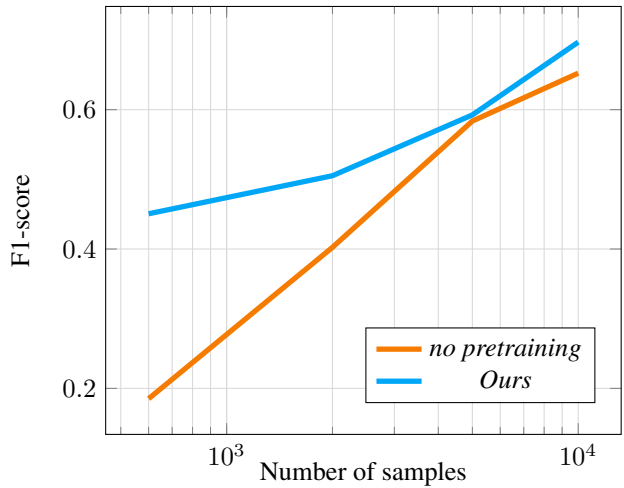
\begin{figure}[ht]
\centering
\begin{tikzpicture}
    \begin{axis}[grid=both,
                 grid style={solid,gray!30!white},
                 xlabel={Number of samples},
                 ylabel={F1-score},
                 x label style={at={(axis description cs:0.5,0.05)},anchor=north},
                 y label style={at={(axis description cs:0.05,.5)},anchor=south},
                 xmode=log,
                 legend style={at={(0.95,0.25)}},
                 ]
        \addplot[line width=2pt, color=tb_color_1] table [x=Step, y=Value, col sep=comma] {figs/dataset_size_ablation_plot/suppl_baseline_perf-curve.csv};
        \addlegendentry{\textit{no pretraining}}
        \addlegendentry{\textit{Ours}}
        \addplot[line width=2pt, color=tb_color_2] table [x=Step, y=Value, col sep=comma] {figs/dataset_size_ablation_plot/suppl_ours_perf-curve.csv};
    \end{axis}
  \end{tikzpicture}
  \caption{F1-scores (y-axis) over different target dataset sizes (x-axis). The F1-score is calculated between the node and edge mAP as described in Section \ref{sec:suppl_eval_metrics}. The orange line depicts the F1-scores of the \textit{no-pretraining} baseline, and the blue line with our contributions, as described in Section \ref{sec:results}. The x-axis is in logarithmic scale. We observe that our contributions are significantly reducing data requirements, especially when data is scarce.}
\label{fig:suppl_dataset_curve}
\end{figure}

Lastly, Figure \ref{fig:suppl_dataset_curve} shows the results of an ablation study on the target dataset size. We plot the harmonic mean of node and edge mAP (see Section \ref{sec:results}) of our method and the \textit{no-pretraining} baseline against the size of the target dataset. We observe that our method consistently outperforms the baseline across all dataset sizes. However, as the number of samples increases, the performance difference between the two methods decreases. This observation is expected because transfer learning becomes less effective (and is also less required) when enough target domain samples are available. Our framework is especially useful if target data is scarce.

\section{Model \& training details}
\label{suppl:model_details}

\begin{table*}[ht]
    \small
    \centering
    \caption{Model details for each experiment. The hyperparameters are the same for each model of the respective experiment. The latent space resolution is controlled via the CNN backbone's stride. It determines the feature size between the backbone and the transformer.}
    \begin{tabular}{l|c|c|c|c|c|c|c|c}
    \toprule
    \multirow{4}{*}{\textbf{Target Set}} & \multicolumn{2}{c|}{\textbf{Backbone}} & \textbf{Latent Space} & \multicolumn{4}{c|}{\textbf{Transformer}} & \textbf{FFN} \\ \cmidrule{2-9} 
     & \textbf{Type} & \textbf{Hid. Dim.} & \textbf{Resolution} & \textbf{Hid. Dim.} & \textbf{\# Lay.} & \multicolumn{1}{c|}{\textbf{\begin{tabular}[c]{@{}c@{}}\# Obj\\ Token\end{tabular}}} & \textbf{\begin{tabular}[c]{@{}c@{}}\# RLN\\ Token\end{tabular}} & \textbf{Hid. Dim.} \\ 
     \midrule
     \midrule
    Agadez & ResNet101 & 512 & Multi-Level & 512 & 3 & \multicolumn{1}{c|}{80} & 2 & 1024 \\
    Munich & ResNet101 & 512 & Multi-Level & 512 & 3 & \multicolumn{1}{c|}{80} & 2 & 1024 \\
    Synthetic OCTA & ResNet101 & 512 & Multi-Level & 512 & 3 & \multicolumn{1}{c|}{80} & 5 & 1024 \\
    OCTA-500 & ResNet101 & 512 & Multi-Level & 512 & 3 & \multicolumn{1}{c|}{80} & 2 & 1024 \\
    Synthetic MRI & SeresNet & 256 & $2 \times 2 \times 2$ & 552 & 4 & \multicolumn{1}{c|}{120} & 2 & 1280 \\
    Whole Brain Vessels & SeresNet & 256 & $2 \times 2 \times 2$ & 552 & 4 & \multicolumn{1}{c|}{120} & 2 & 1280 \\
    \bottomrule
    \end{tabular}
    \label{table:model_hps}
\end{table*}

\begin{table*}[ht]
\small
\centering
\caption{Training details and hyperparameters for each trained model.}
\begin{tabular}{c|c|c|cc|ccccc}
\toprule
\multicolumn{1}{c|}{\multirow{3}{*}{\textbf{Experiment}}} & \multirow{3}{*}{\textbf{Batch Size}} & \multirow{3}{*}{\textbf{Epochs}} & \multicolumn{2}{c|}{\textbf{Learning Rate}} & \multicolumn{4}{c}{\textbf{Loss Coeff.}} \\ 
\cmidrule{4-10} 
\multicolumn{1}{c|}{} & &  & \multicolumn{1}{c|}{\textbf{Backbone}} & \textbf{Transformer} & \multicolumn{1}{c|}{$\pmb{\lambda_{gIoU}}$} & \multicolumn{1}{c|}{$\pmb{\lambda_{cls}}$} & \multicolumn{1}{c|}{$\pmb{\lambda_{reg}}$} & \multicolumn{1}{c|}{$\pmb{\lambda_{Resln}}$} & $\pmb{\lambda_{reg}}$\\ 
\midrule
\midrule
A & 32 & 100 & \multicolumn{1}{c|}{0.00002} & 0.0002 & \multicolumn{1}{c|}{2} & \multicolumn{1}{c|}{3} & \multicolumn{1}{c|}{5} &  \multicolumn{1}{c|}{5} & 1.0\\
B & 32 & 100 & \multicolumn{1}{c|}{0.00002} & 0.0002 & \multicolumn{1}{c|}{2} & \multicolumn{1}{c|}{3} & \multicolumn{1}{c|}{5} &  \multicolumn{1}{c|}{5} & 1.0\\
C & 32 & 100 & \multicolumn{1}{c|}{0.00002} & 0.0002 & \multicolumn{1}{c|}{2} & \multicolumn{1}{c|}{3} & \multicolumn{1}{c|}{5} &  \multicolumn{1}{c|}{5} & 1.0\\
D & 32 & 100 & \multicolumn{1}{c|}{0.00007} & 0.00007 & \multicolumn{1}{c|}{3} & \multicolumn{1}{c|}{4} & \multicolumn{1}{c|}{2} &  \multicolumn{1}{c|}{6} & 0.8\\
E & 32 & 100 & \multicolumn{1}{c|}{0.00007} & 0.00007 & \multicolumn{1}{c|}{3} & \multicolumn{1}{c|}{4} & \multicolumn{1}{c|}{2} &  \multicolumn{1}{c|}{6} & 0.8\\
 \bottomrule
\end{tabular}
\label{table:training_hps}
\end{table*}

To find the optimal hyperparameters, we follow a three-step approach. First, we optimize the model architecture hyperparameters (e.g., model size) with a random weight initialization (i.e., no transfer learning) on the target task. Then, we fix these hyperparameters for the remainder of the optimization process. An overview of the model hyperparameters for each experiment can be found in Table \ref{table:model_hps}. Second, we optimize the training hyperparameters (e.g., learning rate or batch size) for pretraining on the source task with the fixed model architecture hyperparameters from step 1. Third, we use the pretrained model with the best performance on the source task and optimize the training parameters on the target task for each training strategy separately on the validation set. We follow this approach because optimizing the whole pipeline (including pretraining and fine-tuning) in a brute-force manner would require too many resources in terms of computational power and energy consumption. Table \ref{table:training_hps} depicts the training hyperparameters for the target task for all the experiments listed in Section \ref{sec:results}.

\section{Datasets}
\label{suppl:datasets}
In the following, we describe the properties and sampling of our six diverse image datasets and the unlabeled datasets we used for the self-supervised baseline. 

\subsection{Training set - 20 U.S. Cities}
\noindent \cite{he2020sat2graph} is a city-scale dataset consisting of satellite remote sensing (SRS) images from 20 U.S. cities and their road graphs covering a total area of 720 km$^2$. The satellite images are retrieved in the RGB format via the Google Maps API \cite{googleapi}. The corresponding road network graphs are extracted from OpenStreetMap \cite{haklay2008osm}. We cut the resulting images and labels into overlapping patches of 128x128 pixels with a spatial resolution of one meter per pixel. In these patches, we eliminate redundant nodes (i.e., nodes of degree 2 with a curvature of fewer than 160 degrees) to simplify the prediction task \cite{belli2019nodeprune}.

\begin{table*}[ht]
\footnotesize
\centering
\caption[Dataset Summary]{Overview of the used datasets, selected characteristics, and respective training, validation, and test set sizes.}
\begin{tabular}{ l | c | c | c | c | c | c | c | c | c }
    \toprule
    \multicolumn{1}{c |}{\multirow{2}{*}{Dataset}} & \multicolumn{4}{c |}{Road Description} & \multicolumn{2}{c |}{Vessel Description}   & \multicolumn{3}{c}{Split} \\
    & Street Type & Vegetation & Layout & Continent &   Dimension & Spatial Size & Train & Val & Test \\  
    \midrule
    20 U.S. Cities \cite{he2020sat2graph} & Sealed & Rich & Grid-plan & N. America & 2D & 128$\times$128 &  99.2k & 24.8k & 25k \\
    \midrule
    Global Diverse Cities & \multicolumn{4}{c |}{}  \\
    \quad Agadez & Unsealed & Arid & Grid-plan & Africa & 2D & 128$\times$128 & 480 & 120 & 290 \\
    \quad Munich & Sealed & Rich & Historical & Europe & 2D & 128$\times$128 & 440 & 110 & 220 \\
    \midrule
    Synth. OCTA \cite{menten2022physiology} & - &- & - & - & 2D & 128$\times$128 & 480 & 120 & 2k \\
    OCTA-500 \cite{li2020octa500} & - &- & - & - &2D & 128$\times$128 & 1.6k & 400 & 2.2k \\
    \midrule
    Synth. MRI \cite{schneider2012tissue} & - &- & - & - &3D & 64$\times$ 64 $\times$64 & 4k & 1k & 5k \\
    Microscopy \cite{todorov2020machine} & - &- & - & - &3D & 50$\times$ 50 $\times$50 & 4k & 1k & 1.2k \\
    \midrule
    \midrule
    Unlabeled datasets & \multicolumn{4}{c |}{}  \\
    \quad 20 U.S. Cities & - & - & - & - & 2D & 128$\times$128 & 124k & - & - \\
    \quad Synth. OCTA \cite{menten2022physiology} & - & - & - & - & 2D & 128$\times$128 & 96.4k & - & - \\
    \quad Real OCTA \cite{li2020octa500, ma2021rose} & - & - & - & - & 2D & 128$\times$128 & 40k & - & - \\
    \quad Synthetic MRI \cite{schneider2012tissue} & - & - & - & - & 3D & 64$\times$ 64 $\times$64 & 80k & - & - \\
    \quad Microscopy \cite{erturk2012three} & - & - & - & - & 3D & 50$\times$ 50 $\times$50 & 43.5k & - & - \\
    
    \bottomrule
\end{tabular}
\label{table:datasets}
\end{table*}

\subsection{Agadez and Munich, cities around the globe}

\noindent We create our own image dataset from OpenStreetMap\footnote{\url{https://www.openstreetmap.org}} covering areas that differ from those covered by the 20 U.S. cities dataset in terms of geographical and structural characteristics. Geographical characteristics refer to the area's natural features (e.g., vegetation), while structural characteristics relate to anthropogenic (human-made) structures that affect an area's surface (e.g., street type) or layout (e.g., city type). The complete dataset contains a 4 km$^2$ area of 11 cities with different characteristics in different parts of the world. Both source images and labels were obtained in the same manner as for the 20 U.S. cities dataset \cite{he2020sat2graph}. Our dataset is accessible in our GitHub repository \footnote{GitHub repository will be made publicly available upon acceptance}.

For our experiments, we choose two cities, Agadez and Munich, whose characteristics differ from the 20 U.S. cities dataset in different aspects as displayed in Table \ref{table:datasets}. We strategically choose those cities to investigate how differences in specific characteristics between the source and target domain affect knowledge transfer and how transfer learning strategies should be adapted to these differences. We especially test our hypothesis that surface-level characteristics are captured by different components than layout-level characteristics. These new datasets enable the verification because Agadez differs from 20 U.S. cities in surface-level characteristics (e.g., vegetation, street type, and buildings) but shares a similar city layout (i.e., grid plan). Note that although Agadez has a historical city center, we chose a part of the city that follows the typical grid layout. Contrary to this, Munich is similar to U.S. cities in surface-level characteristics while following a different city layout (i.e., a historical European city layout). We test each city dataset separately. 

\subsection{Synthetic OCTA}
\noindent The synthetic Optical Coherence Tomography Angiography (OCTA) dataset \cite{menten2022physiology} consists of synthetic OCTA scans with intrinsically matching ground truth labels, namely the corresponding segmentation map and the vessel graphs. The images were created using a simulation based on the physiological principles of angiogenesis to replicate the intricate retinal vascular plexuses \cite{schneider2012tissue}, followed by incorporating physics-based modifications to emulate the image acquisition process of OCTA along with the usual artifacts. We project the 3D OCTA images along the main axis, split them scan-wise between training and testing sets, and extract 2600 overlapping samples of $128 \times 128$ pixels. For training our self-supervised baseline, we use the same procedure on 200 additional synthetic OCTA scans to extract almost 100,000 patches.

\subsection{OCTA-500}
\noindent The OCTA-500 dataset \cite{li2020octa500} includes 300 OCTA scans with a 6 mm $\times$ 6 mm field of view. The $400 \times 400$ large \textit{en-face} projection images were manually annotated with sparse vessel labels. We extract the graphs from these segmentation maps using the method presented by Drees et al. \cite{drees2021voreen}. We split the scans patient-wise between training and testing sets and create around 3000 overlapping patches with a spatial size of $128 \times 128$. Furthermore, we combine the OCTA scans from \textit{OCTA-500} with the scans of the \textit{ROSE} dataset \cite{ma2021rose} to obtain around 40,000 patches for training our self-supervised baseline. This combination is necessary for an unlabeled dataset large enough for self-supervised pretraining.  

\subsection{Synthetic MRI}
\noindent The Synthetic MRI dataset \cite{tetteh2018deepvesselnet} is a synthetical 3D dataset that simulates the characteristics of clinical vessel datasets. The original dataset provides ground truth labels for vessel segmentation, centerlines, and bifurcation points. The ground truth graphs are obtained with the method described by Drees et al. \cite{drees2021voreen}. We cut the volumes and their graphs in overlapping patches of $64 \times 64 \times 64$ voxels. We use the same dataset with all 80,000 patches for our self-supervised pretraining.

\subsection{Whole Brain Vessels}
\noindent The Whole Brain Vessel dataset \cite{paetzold2021whole} is a publicly available open graph benchmark dataset for link prediction (ogbl-vessel\footnote{\url{https://ogb.stanford.edu/docs/linkprop/\#ogbl-vessel}}). It consists of a graph representing the entirety of the mouse brain's vascular structure down to the capillary level. Todorov et al. \cite{todorov2020machine} obtained the raw vessel scans using tissue-clearing methods and fluorescent microscopy and then segmented the brain vasculature using CNNs. The dataset has an image, segmentation, and graph representation. We create overlapping patches with a spatial size of $50 \times 50 \times 50$ voxels and remove artifactual patches (e.g., patches containing only noise). We extract 43,500 image patches from an unlabeled whole-brain mouse scan obtained with the vDISCO pipeline \cite{erturk2012three} for training our self-supervised model.

\section{Evaluation metrics}
\label{sec:suppl_eval_metrics}
We choose to evaluate our models' performance using three different evaluation metric types: 1) topological metrics, 2) graph distance metrics, and 3) object detection metrics.

\paragraph{Topological metrics} The TOPO-score \cite{biagioni2012topo} samples multiple sub-graphs starting from different seed locations from the ground truth and measures its similarity to the inferred graph from the predicted graph with the same seed location. The similarity is measured by matching a fixed amount of points between the two graphs. Two points from two graphs are matched if the distance between their spatial coordinates is below a threshold. The result of this matching across all sampled subgraphs is used for calculating precision and recall. This method accurately quantifies a prediction's geometrical (i.e., the roads' geographical position) and topological (i.e., the roads' interconnections) quality. We use the implementation and parameters from Biagioni et al. \cite{biagioni2012topo}. These metrics are not implemented in 3D.

\paragraph{Graph distance metrics} The street mover distance (SMD) approximates the Wasserstein distance between a fixed number of uniformly sampled points along the ground truth graph and the predicted graph. Intuitively, it represents the minimal distance by which the predicted graph must be moved to match the ground truth \cite{belli2019nodeprune}.

\paragraph{Object detection metrics} Further, we resort to widely-used object detection metrics: mean average precision (mAP) and mean average recall (mAR) \cite{padilla2020detMetrics}. To calculate each detection's intersection over union (IoU), we create a hypothetical bounding box of fixed size around each node. Similarly, we create bounding boxes around the edges with a minimum spatial size $m$ in all dimensions. This minimum holds for edges that connect two nodes $a$ and $b$ where the difference between the coordinates in one dimension is lower than $m$ (e.g. if $|a_x - b_x| < m$). We calculate the mean AP and AR between the values of different IoU thresholds (i.e., 0.5 and 0.95).


\section{Additional quantitative results}
In Table \ref{table:suppl_main_results_std}, we present our main results from Table \ref{table:main_results} in addition to the results' standard deviation across five mutually exclusive folds of the test set. 

\begin{table*}[t]
    \centering
    \footnotesize
    \caption{Main results with standard deviations. Quantitative Results for our cross-dimensional image-to-graph transfer learning framework. All models are pretrained on the U.S cities road dataset. We outperform the baselines across all datasets. We present the standard deviations in addition to the main results.}
    \begin{tabular}{ l | l | c | c | c | c | c | c | c }
        \toprule
        \begin{tabular}[c]{@{}l@{}} \textbf{Fine Tuning} \\ \textbf{Training Set}\end{tabular} 
        & \begin{tabular}[c]{@{}l@{}} \textbf{(Pre-)Training} \\ \textbf{Strategy}\end{tabular} 
        & \begin{tabular}[c]{@{}l@{}} \textbf{Node-}     \\ \textbf{mAP}$\uparrow$\end{tabular} 
        & \begin{tabular}[c]{@{}l@{}} \textbf{Node-}     \\ \textbf{mAR}$\uparrow$\end{tabular} 
        & \begin{tabular}[c]{@{}l@{}} \textbf{Edge-}     \\ \textbf{mAP}$\uparrow$\end{tabular} 
        & \begin{tabular}[c]{@{}l@{}} \textbf{Edge-}     \\ \textbf{mAR}$\uparrow$\end{tabular} 
        & \textbf{SMD $\downarrow$}
            & \begin{tabular}[c]{@{}l@{}} \textbf{Topo-} \\ \textbf{Prec.$\uparrow$ }\end{tabular} 
        & \begin{tabular}[c]{@{}l@{}} \textbf{Topo-} \\ \textbf{Rec.$\uparrow$ }\end{tabular} \\
    
        \midrule
        \midrule 
        \multicolumn{9}{l}{\textbf{A) TL from roads (2D) to roads (2D)} \Tstrut\Bstrut} \\
        \midrule
        \multirow{4}{*}{Agadez \cite{haklay2008osm}}
        &  No Pretr. \cite{glorot2010understanding} & 0.067$\pm$\tiny0.006 & 0.122$\pm$\tiny0.007 & 0.021$\pm$\tiny0.005 & 0.043$\pm$\tiny0.006 & 0.062$\pm$\tiny0.028 & 0.369$\pm$\tiny0.051 & 0.261$\pm$\tiny0.047\\
        &  Self-superv. \cite{chen2021empirical}    & 0.083$\pm$\tiny0.010 & 0.156$\pm$\tiny0.011 & 0.030$\pm$\tiny0.005 & 0.071$\pm$\tiny0.007 & 0.030$\pm$\tiny0.005 & 0.471$\pm$\tiny0.0082 & 0.459$\pm$\tiny0.039\\
        &  Supervised                               & 0.161$\pm$\tiny0.021 & 0.237$\pm$\tiny0.023 & 0.115$\pm$\tiny0.016 & \textbf{0.177$\pm$\tiny0.017} & 0.023$\pm$\tiny0.009 & 0.783$\pm$\tiny0.018 & \textbf{0.711$\pm$\tiny0.039}\\
        &  \textbf{Ours}                            & \textbf{0.163$\pm$\tiny0.017} & \textbf{0.244$\pm$\tiny0.015} & \textbf{0.116$\pm$\tiny0.019} & 0.172$\pm$\tiny0.021 & \textbf{0.022$\pm$\tiny0.003} & \textbf{0.816$\pm$\tiny0.032} & 0.614$\pm$\tiny0.036\\
        
        \midrule
        
        \multirow{4}{*}{Munich \cite{haklay2008osm}}
        &  No Pretr. \cite{glorot2010understanding} & 0.083$\pm$\tiny0.012 & 0.120$\pm$\tiny0.011 & 0.034$\pm$\tiny0.013 & 0.054$\pm$\tiny0.016 & 0.235$\pm$\tiny0.049 & 0.260$\pm$\tiny0.057 & 0.247$\pm$\tiny0.070\\
        &  Self-superv. \cite{chen2021empirical}    & 0.088$\pm$\tiny0.021 & 0.145$\pm$\tiny0.033 & 0.060$\pm$\tiny0.015 & 0.097$\pm$\tiny0.023 & 0.155$\pm$\tiny0.032 & 0.339$\pm$\tiny0.035 & 0.384$\pm$\tiny0.075\\
        &  Supervised                               & 0.277$\pm$\tiny0.022 & 0.336$\pm$\tiny0.025 & 0.207$\pm$\tiny0.027 & 0.272$\pm$\tiny0.031 & 0.091$\pm$\tiny0.038 & 0.682$\pm$\tiny0.037 & \textbf{0.660$\pm$\tiny0.041}\\
        &  \textbf{Ours}                            & \textbf{0.285$\pm$\tiny0.015} & \textbf{0.344$\pm$\tiny0.011} & \textbf{0.224$\pm$\tiny0.030} & \textbf{0.277$\pm$\tiny0.031} & \textbf{0.090$\pm$\tiny0.043} & \textbf{0.726$\pm$\tiny0.078} & 0.655$\pm$\tiny0.070\\
        \midrule
        \midrule
        
        \multicolumn{9}{l}{\textbf{B) TL from roads (2D) to synthetic retinal vessels (2D) }\Tstrut\Bstrut}  \\
        \midrule
        \multirow{4}{*}{\begin{tabular}[c]{@{}l@{}} Synthetic \\ OCTA \cite{menten2022physiology}\end{tabular}}
        &  No Pretr. \cite{glorot2010understanding} & 0.273$\pm$\tiny0.003 & 0.375$\pm$\tiny0.003 & 0.140$\pm$\tiny0.002 & 0.339$\pm$\tiny0.003 & 0.005$\pm$\tiny0.002 & 0.181$\pm$\tiny0.004 & 0.948$\pm$\tiny0.004\\
        &  Self-superv. \cite{chen2021empirical}    & 0.136$\pm$\tiny0.002 & 0.260$\pm$\tiny0.003 & 0.069$\pm$\tiny0.002 & 0.223$\pm$\tiny0.004 & 0.031$\pm$\tiny0.006 & 0.093$\pm$\tiny0.005 & 0.927$\pm$\tiny0.010\\
        &  Supervised                               & 0.291$\pm$\tiny0.003 & 0.384$\pm$\tiny0.003 & 0.170$\pm$\tiny0.002 & 0.338$\pm$\tiny0.005 & 0.004$\pm$\tiny0.001 & 0.211$\pm$\tiny0.005 & \textbf{0.957$\pm$\tiny0.007}\\
        &  \textbf{Ours}                            & \textbf{0.415$\pm$\tiny0.005} & \textbf{0.493$\pm$\tiny0.003} & \textbf{0.250$\pm$\tiny0.004} & \textbf{0.415$\pm$\tiny0.004} & \textbf{0.002$\pm$\tiny0.001} & \textbf{0.401$\pm$\tiny0.003} & 0.890$\pm$\tiny0.007\\
        \midrule
        \midrule
        
        \multicolumn{9}{l}{\textbf{C) TL from roads (2D) to real retinal vessels (2D) }\Tstrut\Bstrut}  \\
        \midrule
        \multirow{4}{*}{OCTA-500\cite{li2020octa500}}
        &  No Pretr. \cite{glorot2010understanding} & 0.189$\pm$\tiny0.005 & 0.282$\pm$\tiny0.007 & 0.108$\pm$\tiny0.004 & 0.169$\pm$\tiny0.006 & 0.017$\pm$\tiny0.002 & 0.737$\pm$\tiny0.007 & 0.634$\pm$\tiny0.010\\
        &  Self-superv. \cite{chen2021empirical}    & 0.214$\pm$\tiny0.004 & 0.305$\pm$\tiny0.004 & 0.135$\pm$\tiny0.001 & 0.213$\pm$\tiny0.002 & 0.016$\pm$\tiny0.002 & 0.763$\pm$\tiny0.012 & 0.706$\pm$\tiny0.005\\
        &  Supervised                               & 0.366$\pm$\tiny0.004 & 0.447$\pm$\tiny0.004 & 0.276$\pm$\tiny0.006 & 0.354$\pm$\tiny0.007 & 0.014$\pm$\tiny0.001 & 0.862$\pm$\tiny0.010 & 0.775$\pm$\tiny0.011\\
        &  \textbf{Ours}                            & \textbf{0.491$\pm$\tiny0.006} & \textbf{0.571$\pm$\tiny0.005} & \textbf{0.366$\pm$\tiny0.009} & \textbf{0.489$\pm$\tiny0.007} & \textbf{0.012$\pm$\tiny0.002} & \textbf{0.877$\pm$\tiny0.004} & \textbf{0.817$\pm$\tiny0.011}\\
        \midrule
        \midrule
        
        \multicolumn{9}{l}{\textbf{D) TL from roads (2D) to brain vessels (3D)}\Tstrut\Bstrut} \\
        \midrule
        \multirow{4}{*}{\begin{tabular}[c]{@{}l@{}} Synthetic \\ MRI \cite{schneider2012tissue}\end{tabular}}
        &  No Pretr. \cite{glorot2010understanding} & 0.162$\pm$\tiny0.003 & 0.250$\pm$\tiny0.003 & 0.125$\pm$\tiny0.004 & 0.201$\pm$\tiny0.004 & \textbf{0.013$\pm$\tiny0.000} & - & -\\
        &  Self-superv. \cite{chen2021empirical}    & 0.162$\pm$\tiny0.003 & 0.252$\pm$\tiny0.003 & 0.120$\pm$\tiny0.004 & 0.193$\pm$\tiny0.004 & 0.014$\pm$\tiny0.000 & - & -\\
        &  Supervised                               & $\star$ & $\star$ & $\star$  & $\star$  & $\star$  & $\star$ & $\star$\\
        &  \textbf{Ours}                            & \textbf{0.356$\pm$\tiny0.003} & \textbf{0.450$\pm$\tiny0.002} & \textbf{0.221$\pm$\tiny0.003} & \textbf{0.322$\pm$\tiny0.003} & \textbf{0.013$\pm$\tiny0.000} &  - & -\\
        \midrule
        \midrule
        \multicolumn{9}{l}{\textbf{E) TL from roads (2D) to real whole-brain vessel data (3D)}\Tstrut\Bstrut} \\
        \midrule
        \multirow{4}{*}{\begin{tabular}[c]{@{}l@{}} Microscopic \\ images \cite{todorov2020machine}\end{tabular}}
        &  No Pretr. \cite{glorot2010understanding} & 0.231$\pm$\tiny0.016 & 0.308$\pm$\tiny0.021 & 0.249$\pm$\tiny0.017 & 0.329$\pm$\tiny0.023 & \textbf{0.017$\pm$\tiny0.000} & - & -\\
        &  Self-superv. \cite{chen2021empirical}    & 0.344$\pm$\tiny0.026 & 0.404$\pm$\tiny0.029 & 0.363$\pm$\tiny0.026 & 0.425$\pm$\tiny0.030 & \textbf{0.017$\pm$\tiny0.000} & - & -\\
        &  Supervised                               & $\star$  & $\star$  & $\star$  & $\star$  & $\star$  & $\star$ & $\star$\\
        &  \textbf{Ours}                            & \textbf{0.483$\pm$\tiny0.037} & \textbf{0.535$\pm$\tiny0.039} & \textbf{0.523$\pm$\tiny0.041} & \textbf{0.566$\pm$\tiny0.043} & \textbf{0.017$\pm$\tiny0.000} & - & -\\
        \bottomrule
    \end{tabular}
    \vspace{0.15cm}
    \label{table:suppl_main_results_std}
    \end{table*}

\section{Additional qualitative results }
\begin{figure*}[ht!]
\centerline{\includegraphics[width=0.99\linewidth]{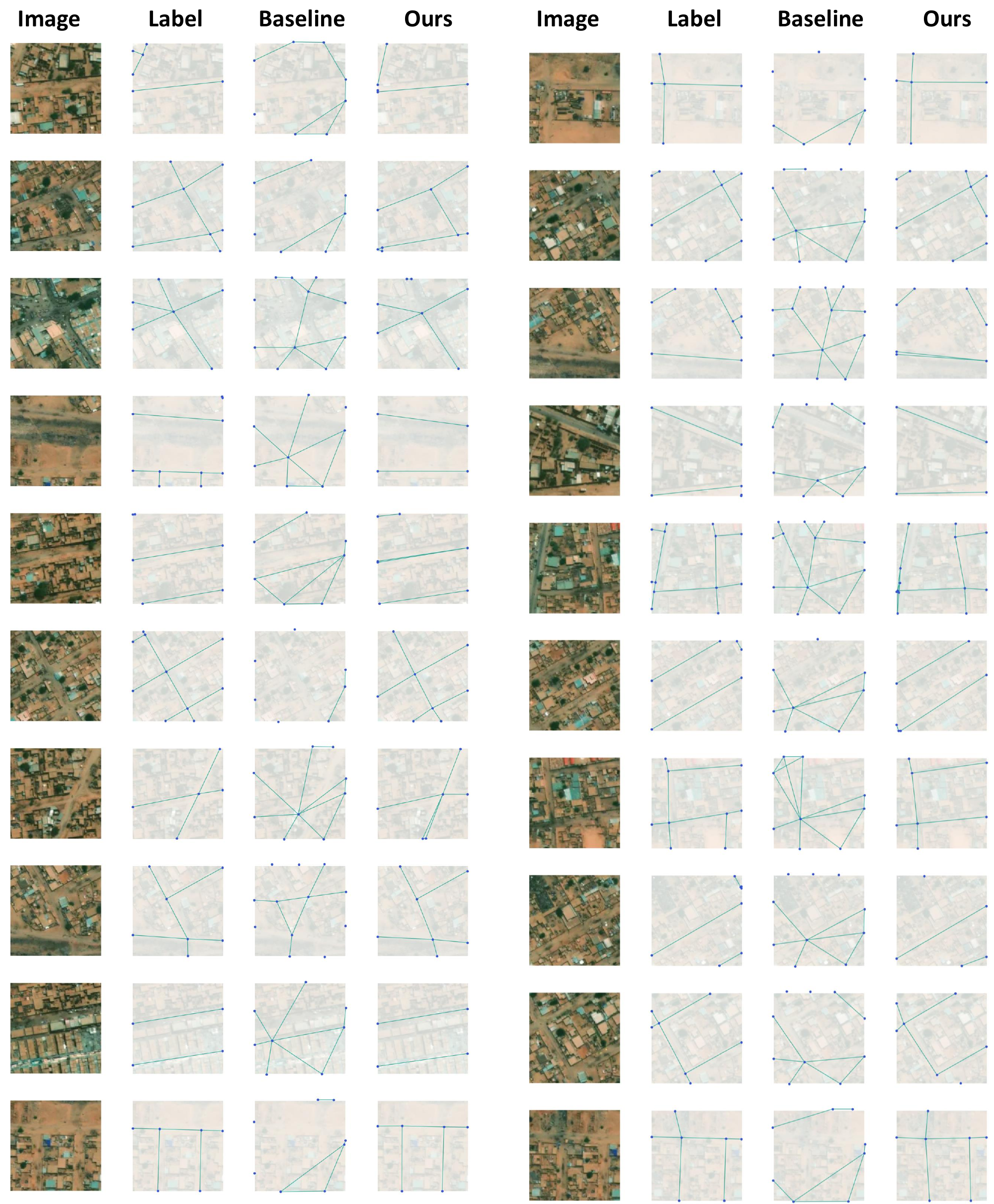}}
\caption{Qualitative results for the Agadez dataset. Two columns, from left to right: Image, ground truth graph, baseline, and our method. Our method consistently outperforms the baselines, which overpredict the edges and nodes for road data. }
\label{fig:suppl_agadez}
\end{figure*}

\begin{figure*}[ht!]
\centerline{\includegraphics[width=0.99\linewidth]{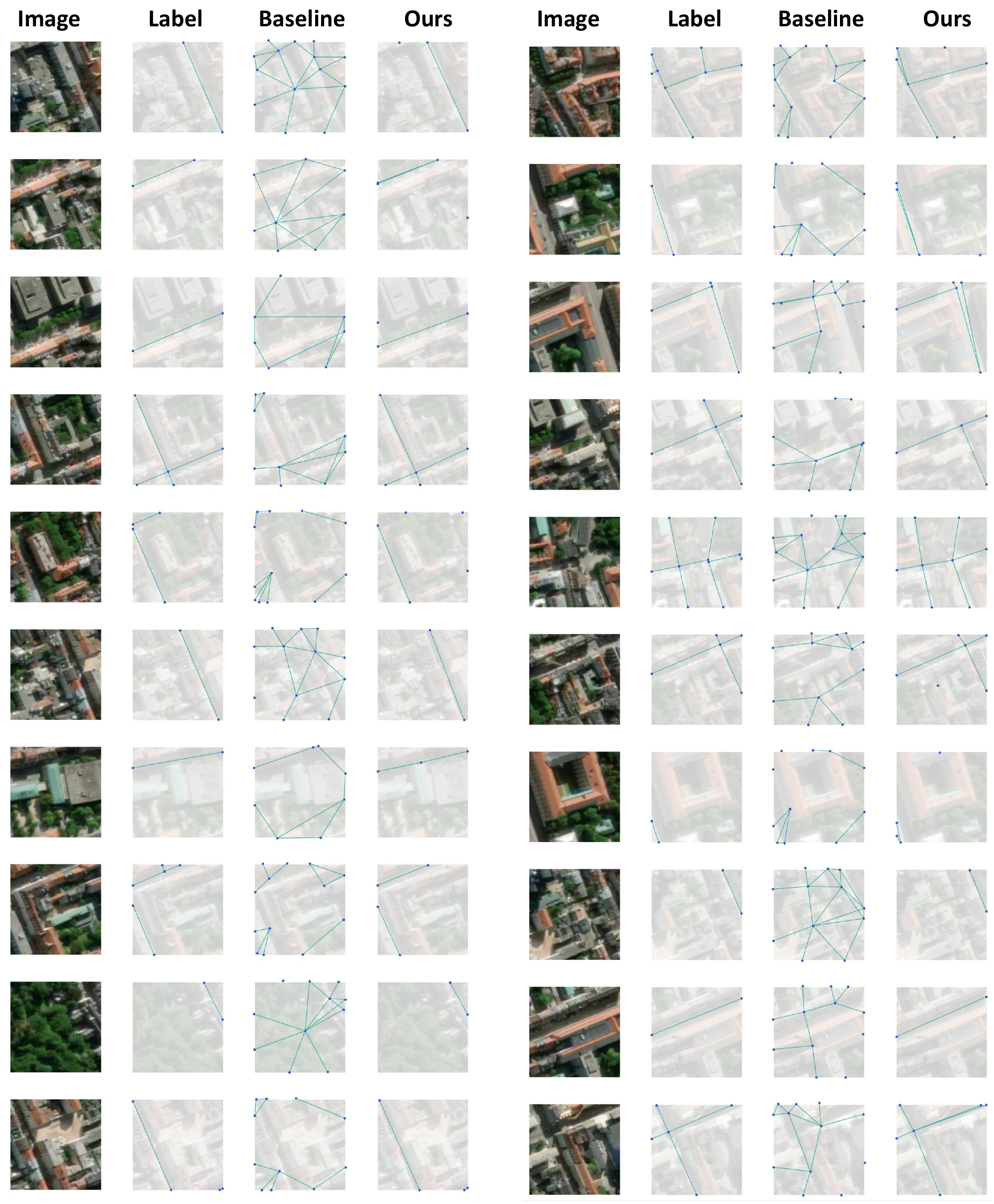}}
\caption{Qualitative results for the Munich dataset. Two columns, from left to right: Image, ground truth graph, baseline, and our method. Our method consistently outperforms the baselines, which overpredict the edges and nodes for road data. }
\label{fig:suppl_munich}
\end{figure*}

\begin{figure*}[ht!]
\centerline{\includegraphics[width=0.99\linewidth]{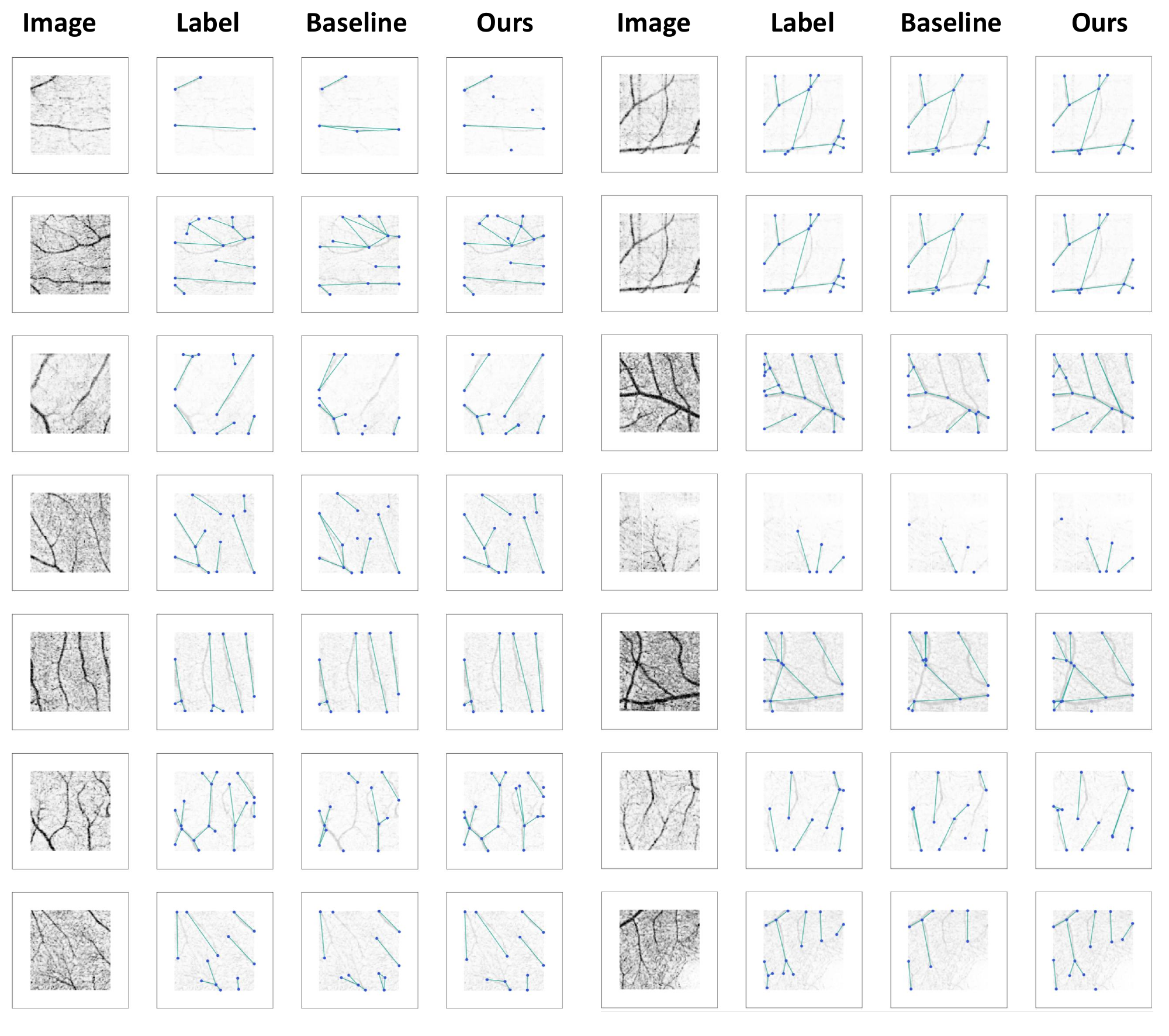}}
\caption{Qualitative results for the OCTA-500 dataset. Two columns, from left to right: Image, ground truth graph, baseline, and our method. Our method consistently outperforms the baselines, which underpredict the edges and nodes for the vessel data. It is important to note that the OCTA-500 dataset labels are on the large vessels. The graph annotations are not provided for all capillaries and are therefore not learned by the models either.}
\label{fig:suppl_2D_real}
\end{figure*}

\begin{figure*}[ht!]
\centerline{\includegraphics[width=0.99\linewidth]{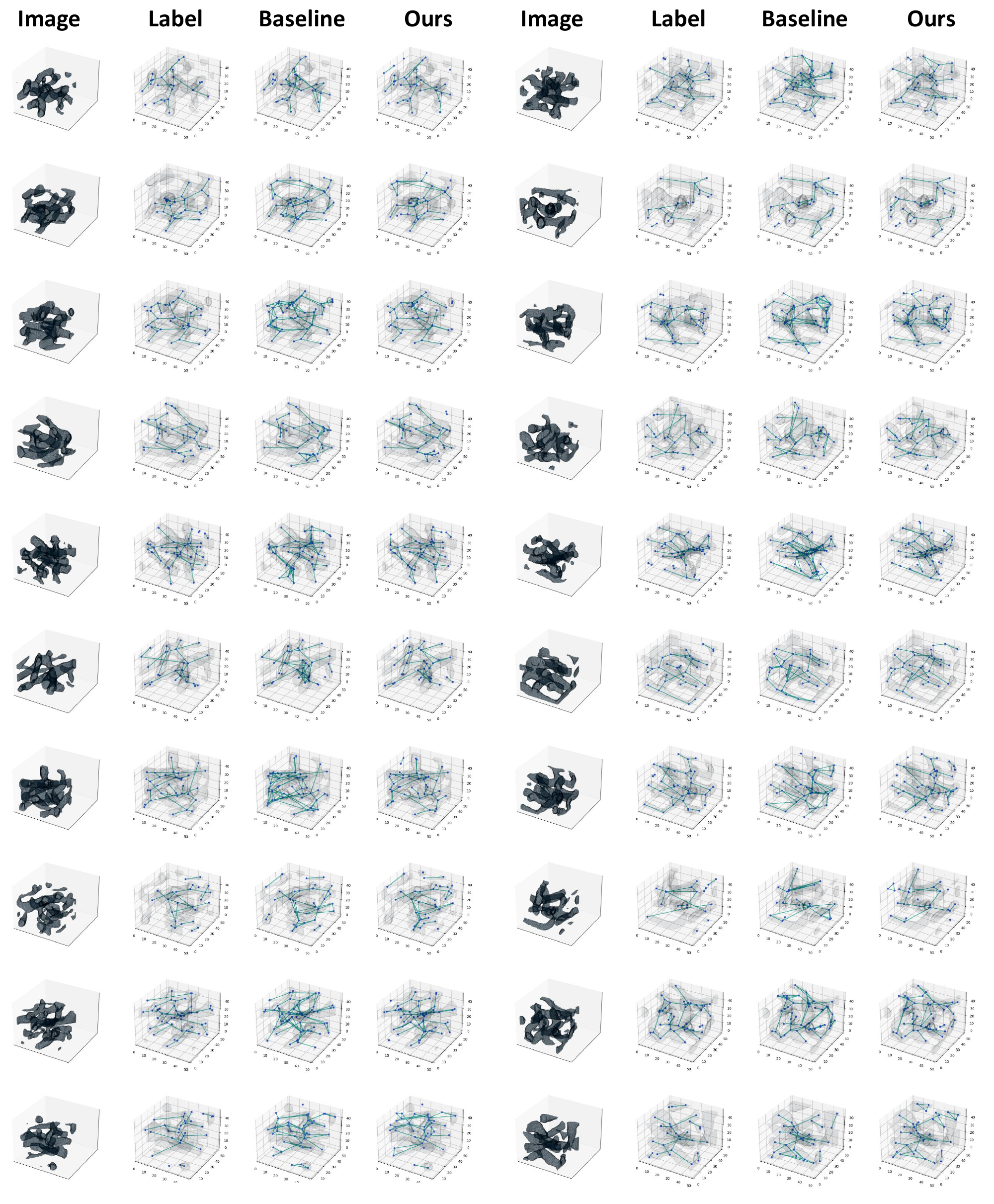}}
\caption{Qualitative results for the 3D whole brain vessel dataset. Two columns, from left to right: Image, ground truth graph, baseline, and our method. Our method consistently outperforms the baselines, which overpredict the edges for the 3D vessel data. Furthermore, the baseline often predicts implausible triangles between three nodes. }
\label{fig:suppl_3D_real}
\end{figure*}

\begin{figure*}[ht!]
\centerline{\includegraphics[width=0.99\linewidth]{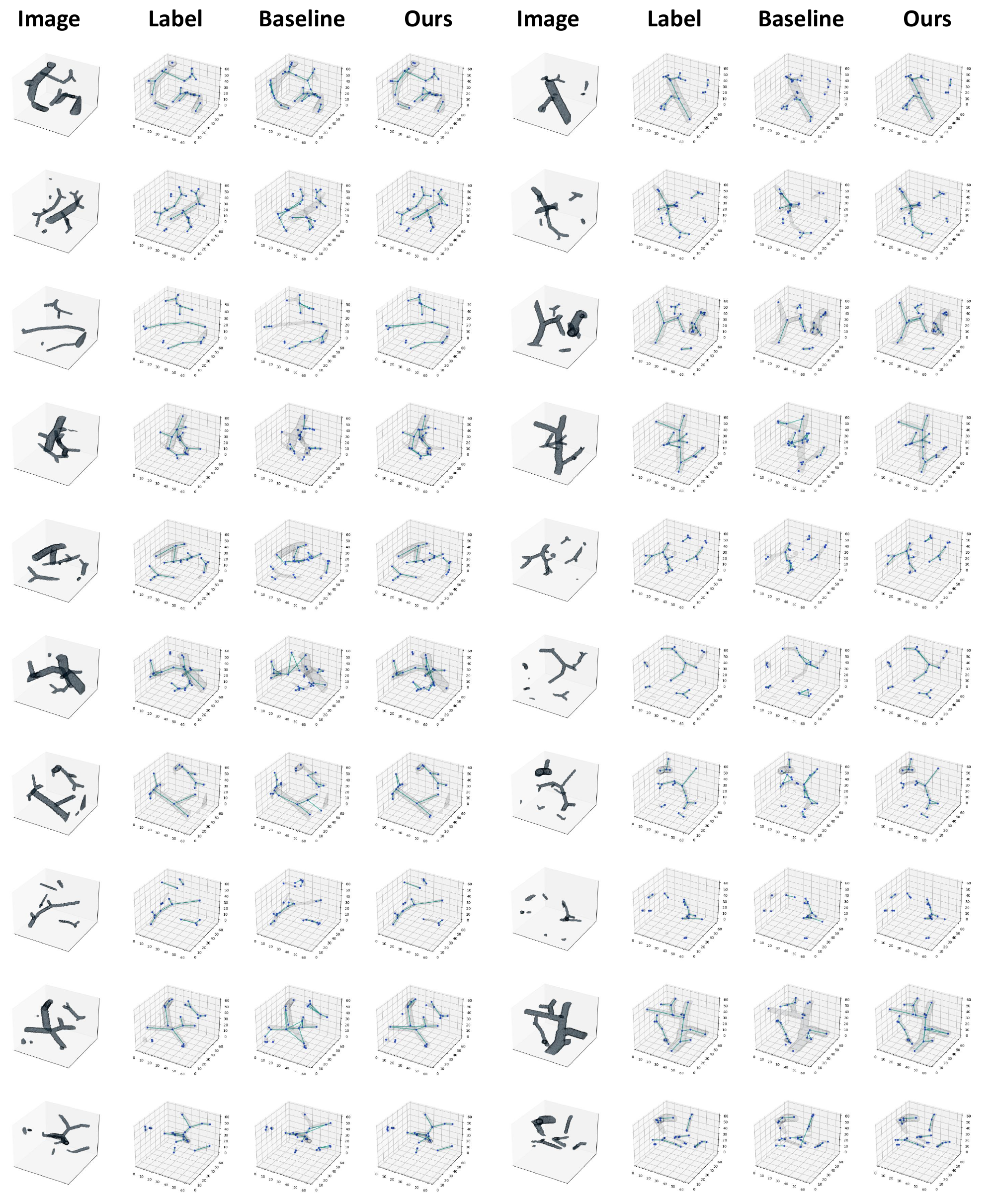}}
\caption{Qualitative results for the synthetic 3D vessel MRI dataset. Two columns, from left to right: Image, ground truth graph, baseline, and our method. Our method consistently outperforms the baselines, which overpredict the nodes for the 3D vessel data and underpredict edges.}
\label{fig:suppl_3D_synth}
\end{figure*}

\noindent We are providing additional qualitative results in the form of multiple figures; please see Figure \ref{fig:suppl_agadez} - \ref{fig:suppl_3D_synth}.
\label{suppl:qual_res}

\clearpage

\end{document}